\documentclass[letterpaper,twocolumn,10pt]{article}
\usepackage{usenix-2020-09}

\usepackage{tikz}
\usepackage{amsmath}

\usepackage[utf8]{inputenc} 
\usepackage[T1]{fontenc}    
\usepackage{hyperref}       
\usepackage{url}            
\usepackage{booktabs}       
\usepackage{amsfonts}       
\usepackage{nicefrac}       
\usepackage{microtype}      
\usepackage{xcolor}         
\usepackage[ruled, lined, linesnumbered, commentsnumbered, longend]{algorithm2e}
\usepackage{dsfont}
\usepackage{array,multirow}
\usepackage{graphicx}
\usepackage{subcaption}
 
\usepackage{color}
\usepackage{xspace}
\usepackage{url}
\usepackage{pifont} 

\usepackage{silence} 
\usepackage[normalem]{ulem} 

\usepackage{enumitem} 
\setlist{nosep} 

\usepackage{wrapfig}
\usepackage{rotating} 

\usepackage{booktabs} 

\usepackage{array} 
\newcommand{\PreserveBackslash}[1]{\let\temp=\\#1\let\\=\temp}
\newcolumntype{C}[1]{>{\PreserveBackslash\centering}p{#1}}
\newcolumntype{R}[1]{>{\PreserveBackslash\raggedleft}p{#1}}
\newcolumntype{L}[1]{>{\PreserveBackslash\raggedright}p{#1}}

\newcommand{\syslong}[0]{Robust Representation Matching\xspace}
\newcommand{\sys}[0]{RRM\xspace}

\newcommand{\cifar}[0]{CIFAR-10\xspace}
\newcommand{\imagenet}[0]{ImageNet\xspace}
\newcommand{\rimagenet}[0]{Restricted-ImageNet\xspace}

\newcommand{\alexnet}[0]{AlexNet\xspace}
\newcommand{\vggs}[0]{VGG11\xspace}
\newcommand{\vggm}[0]{VGG16\xspace}
\newcommand{\vggl}[0]{VGG19\xspace}
\newcommand{\resnets}[0]{ResNet18\xspace}
\newcommand{\resnetl}[0]{ResNet50\xspace}
\newcommand{\github}[0]{\url{https://github.com/Ethos-lab/robust-representation-matching/releases/tag/final}}

\newif\ifsubmit
\submitfalse

\ifsubmit
    \newcommand{\amir}[1]{}
    \newcommand{\kevin}[1]{}
    \newcommand{\pratik}[1]{}
    
    \newcommand{\todo}[1]{}
    
\else
    
    \newcommand{\amir}[1]{\textcolor{blue}{Amir: #1}}
    \newcommand{\kevin}[1]{\textcolor{green}{Kevin: #1}}
    \newcommand{\pratik}[1]{\textcolor{magenta}{Pratik: #1}}
    
    \newcommand{\todo}[1]{\textcolor{red}{TODO: #1}}
   
\fi

\newcommand{\paratitle}[1]{\noindent\textbf{\textit{#1.}}\xspace}

\newcommand{\ie}[0]{\emph{i.e.,}\xspace}
\newcommand{\etal}[0]{\emph{et~al.}\xspace}
\newcommand{\eg}[0]{\emph{e.g.,}\xspace}
\newcommand{\etc}[0]{\emph{etc.}\xspace}

\usepackage{cleveref} 

\crefformat{section}{\S#2#1#3} 
\crefformat{subsection}{\S#2#1#3}
\crefformat{subsubsection}{\S#2#1#3}

\SetKwInput{KwInput}{Input}
\SetKwInput{KwOutput}{Output}

\usepackage[available,functional,reproduced]{usenixbadges}

\begin{document}

\date{}

\title{Transferring Adversarial Robustness\\ Through \syslong}

\author{
{\rm Pratik Vaishnavi}\\
Stony Brook University\\
pvaishnavi@cs.stonybrook.edu
\and
{\rm Kevin Eykholt}\\
IBM\\
kheykholt@ibm.com
\and
{\rm Amir Rahmati}\\
Stony Brook University\\
amir@cs.stonybrook.edu
} 

\maketitle

\begin{abstract}
    With the widespread use of machine learning, concerns over its security and reliability have become prevalent. As such, many have developed defenses to harden neural networks against adversarial examples, imperceptibly perturbed inputs that are reliably misclassified. 
    Adversarial training in which adversarial examples are generated and used during training is one of the few known defenses able to reliably withstand such attacks against neural networks. However, adversarial training imposes a significant training overhead and scales poorly with model complexity and input dimension.
    In this paper, we propose \textit{\syslong (\sys)}, a low-cost method to transfer the robustness of an adversarially trained model to a new model being trained for the same task irrespective of architectural differences. Inspired by student-teacher learning, our method introduces a novel training loss that encourages the student to learn the teacher's robust representations. Compared to prior works, \sys is superior with respect to both model performance and adversarial training time. On \cifar, \sys trains a robust model $\sim1.8\times$ faster than the state-of-the-art. Furthermore, \sys remains effective on higher-dimensional datasets. On \rimagenet, \sys trains a \resnetl model $\sim18\times$ faster than standard adversarial training.

\end{abstract}

\section{Introduction}
    Despite state-of-the-art performance in numerous domains, deep neural networks (DNNs) remain vulnerable to adversarial examples, inputs that are imperceptibly modified such that they are misclassified by DNNs~\cite{szegedy2014intriguing}. In response to the discovery of adversarial examples, several techniques have been proposed to improve the robustness of DNNs against such inputs~\cite{madry2018towards,papernot2016distillation,wong2018provable}. Adversarial training is one such technique that augments the training data with adversarial examples. During training, adversarial examples are generated on the fly and used to tune the network weights. Although adversarial training is simple to implement and secure against a wide array of attacks~\cite{athalye2018obfuscated,madry2018towards}, it slows down the training process significantly and scales poorly with respect to model complexity and input dimension. In our experiments, for example, adversarial training is on average $\sim7\times$ slower than natural training. The expensive computational cost of adversarial training is only exacerbated by improvements to model architecture or new data. When new state-of-the-art model architectures are developed, adversarial training must be re-done in order to obtain adversarially robust models. These events make adversarial training impractical to use in real world settings, where models are frequently tweaked to improve performance. Therefore, it is desirable to be able to transfer adversarial robustness between models of different architectures to reduce the cost associated with adversarial training.
    
    Ilyas~\etal~\cite{Ilyas2019AdversarialEA} demonstrated that adversarial examples are the result of a model's reliance on non-robust features, \ie highly predictive features that are incomprehensible to humans, whose correlation with the predicted label can be easily flipped with a small amount of noise. They argue that adversarial training works by forcing the model to assign higher priority to the robust features for classification. Thus, if the non-robust information can be removed from the dataset, adversarially robust models should be obtainable through standard training.  To this end, they design a \textit{robust dataset} generation process, which first adversarially trains a model and then uses the learned features in the model to transform the original dataset. Their results demonstrated that new models naturally trained on the robust dataset were more adversarially robust than the models naturally trained on the original dataset.

    While the work by Ilyas~\etal~\cite{Ilyas2019AdversarialEA} is a step towards transferable adversarial robustness, it suffers from two significant limitations. First, compared to an adversarially trained model, the adversarial robustness of a model trained on the robust dataset is poor. When evaluated against $\ell_2$-bounded adversary with $\epsilon=0.5$, the adversarial constraint, a \resnetl model trained on the robust dataset achieves $21.8\%$ adversarial accuracy. This is a significant improvement over the same model trained on original dataset, which exhibits $0\%$ adversarial accuracy. However, the performance falls significantly when evaluated against $\epsilon=1.0$, the value used for adversarial training. In this case, the \resnetl model trained on robust data achieves only $2.3\%$ adversarial accuracy. Second, generating a robust dataset from an existing adversarially trained model is slow. In our experiments, it took approximately $6$ hours with a Titan V GPU to generate a robust \cifar dataset using an adversarially trained \resnets model and the default hyperparameters provided by Ilyas~\etal~\cite{Ilyas2019AdversarialEA}. 

    In this paper, we propose \syslong (\sys), a novel, low-cost method to transfer adversarial robustness using a student-teacher framework. Similar to prior works, \sys first adversarially trains a model. Then, using the adversarially trained model as a teacher, \sys trains a new student model by modifying the training loss to include a novel \textbf{robust representation loss} term. This new term encourages the student model to learn the teacher's robust features. In essence, \sys transfers the teacher's robust features directly to the student as part of standard training rather than expecting the student to learn robust features from the data. \sys can transfer robustness even between models of different architectures. Our proposed method outperforms other adversarial robustness transfer methods including the method demonstrated by Goldblum~\etal~\cite{Goldblum_2020}, which employs a more traditional distillation approach~\cite{hinton2015distilling} to transfer adversarial robustness from an adversarially trained teacher to a student using standard training.

    Using \sys's student-teacher paradigm significantly speeds up the process of training adversarially robust models. To demonstrate this, we compare it against other approaches that speed up adversarial training~\cite{shafahi2019adversarial,wong2019fast} using \cifar. Given a pre-trained teacher, \sys is able to achieve performance comparable to adversarial training~\cite{madry2018towards} in the least amount of training time. Additionally, we demonstrate that \sys can benefit from techniques that speed up adversarial training such as Fast Adversarial Training~\cite{wong2019fast}. When combined with Fast Adversarial training, we show \sys achieves higher performance ($\sim \mathbf{2.5\%}$ higher standard accuracy and $\sim \mathbf{3.5\%}$ higher adversarial accuracy) while requiring significantly lower total training time (converges $\sim \mathbf{1.8\times}$ faster) compared to Free Adversarial Training~\cite{shafahi2019adversarial}. Furthermore, we  show that \sys is able to scale to higher dimensional datasets using the \rimagenet dataset. In the presence of an adversarially trained \alexnet model, we are able to train a \resnetl model $\sim18\times$ faster than adversarial training while achieving competitive performance on both natural and adversarial images.


    \paratitle{Our contributions}
    \begin{itemize}
        \setlength\itemsep{1em}
        \item We introduce \syslong (\sys), a technique that allows for transfer of adversarial robustness between two models of varying architecture.
        \item We evaluate \sys on the \cifar and \rimagenet datasets. On \cifar, \sys is able to achieve performance comparable to adversarial training in least amount of training time compared to prior works.
        \item We show that \sys can scale to higher dimensional datasets such as \rimagenet. Compared to adversarial training, \sys trains a robust model $\sim18\times$ faster with a modest reduction in performance ($\sim6\%$ on natural images and $\sim12\%$ on adversarial images).
    \end{itemize}

\section{Background}
\label{sec:background}
In this section, we formally define the problem of adversarial robustness and briefly discuss the concepts foundational to our solution's design.
    \subsection{Preliminaries}
    
        In this paper, we focus on the DNN-based image classification models. The process for training a $C$-class image classifier $F$ parameterized by $\theta$, \ie $F_{\theta}: \mathds{R}^d \rightarrow \{1 \cdots C\}$, involves updating $\theta$ so as to minimize the empirical risk over image $x \in \mathds{R}^d$ and label $y\in\{1,\cdots,C\}$ pairs sampled from an underlying data distribution $\mathcal{D}$. This process, referred to as Empirical Risk Minimization (ERM), can be formalized as follows:
        
        \begin{equation} \label{eqn:stderm}
        \min_\theta \mathds{E}_{(x,y) \sim \mathcal{D}}[\mathcal{L}(F_{\theta}(x), y)]
        \end{equation}

        Here $\mathcal{L}$ is a loss function suitable for the task at hand. For image classification, the cross-entropy loss function is typically used for this purpose. In adversarial machine learning literature, training using the ERM objective is popularly referred to as \textbf{natural} or \textbf{standard training}.

    \subsection{Adversarial Evasion Attacks}\label{sec:evasionattacks}
        Recent literature has exposed several previously unknown vulnerabilities associated with DNN-based image classifiers~\cite{chakraborty2018adversarial}. One such class of vulnerabilities, called \textit{adversarial evasion attacks}, tries to compute \textit{imperceptible} perturbations to the input such that the perturbed input is misclassified by a classifier~\cite{szegedy2014intriguing,goodfellow2014explaining}. Since their discovery, several attacks have been proposed. The most powerful class of attacks uses the first-order gradients of the classifier to compute the necessary perturbations to cause misclassification. The optimization objective for adversarially perturbing a given image $x$, \ie adversary's objective, can be formalized as follows:

        \begin{equation} \label{eqn:advrisk}
        \max_{\delta: \hspace{0.2em} d(x+\delta, x) \leq \epsilon} \mathcal{L}(F_{\theta}(x + \delta), y)
        \end{equation}

       Here, $\delta$ represents the adversarial perturbation. A distance function $d$ and a scalar $\epsilon$ are used to define the set of all permissible adversarial perturbations (or adversary's budget). With respect to images, this is also referred to as the \textit{imperceptibility condition} and it used to ensure that the adversary perturbs the image imperceptibly in order to launch a stealthy attack. The imperceptibility condition for images is often defined using $\ell_{p}$-norm: $||\delta||_p \leq \epsilon$. Note that most gradient-based attacks assume white-box access to the classifier, \ie the same access as the entity that trained the classifier. Some notable gradient-based attacks are JSMA~\cite{papernot2015limitations}, PGD~\cite{madry2018towards}, and CW~\cite{carlini2017evaluating}.

    \subsection{Defending against Evasion Attacks}
        To mitigate the security risks associated with evasion attacks, several defenses have been proposed in recent literature~\cite{ papernot2016distillation,madry2018towards, wong2018provable,cohen2019certified,zhang2019stable}. Papernot~\etal~\cite{papernot2016distillation} used knowledge distillation~\cite{hinton2015distilling} to train image classifiers that are robust against evasion attacks in a process they called defensive distillation. It was later shown that defensive distillation was effective only because the proposed method caused the gradients to vanish, making it difficult for the adversary to find a solution for Equation~\ref{eqn:advrisk}. Through proper scaling of the classifier's outputs, Carlini~and~Wagner~\cite{carlini2016defensive} were able to resolve the issue of vanishing gradients caused by defensive distillation, allowing existing attacks to converge to a solution successfully.
        
        Several subsequent defenses faced similar issues as defensive distillation as all of these works relied on some form of \textit{gradient obfuscation}. Gradient obfuscation prevents an attacker from using the gradient in order to find a solution for Equation~\ref{eqn:advrisk}. However, this approach only serves to protect a model against naive attackers. Adaptive attackers aware of gradient obfuscation can use alternative methods to approximate the gradients and circumvent the defense. In their paper, Athalye~\etal~\cite{athalye2018obfuscated} proposed a general optimization strategy for breaking gradient obfuscation defenses and demonstrated its effectiveness on several published defenses. 
        
        One popular defense that has stood the test of time is adversarial training, first proposed by Madry~\etal~\cite{madry2018towards}, which modifies the training process to create adversarially robust models. Recently, similar training modification defense strategies have been proposed, but with a focus on establishing mathematically provable guarantees~\cite{wong2018provable,cohen2019certified,zhang2019stable} of performance in adversarial environments.
    
        \subsubsection{Adversarial Training}
            The traditional ERM objective (see Equation~\ref{eqn:stderm}) only optimizes for performance in the standard scenario where the data distribution during testing closely resembles the training distribution $\mathcal{D}$. Therefore, an image classifier trained using ERM can not be expected to perform well on adversarial inputs as these inputs deviate significantly from the distribution $\mathcal{D}$. Madry~\etal~\cite{madry2018towards} recognize this drawback of ERM and propose modifications to it that enables training of adversarially robust image classifiers. Instead of minimizing the risk over examples drawn from $\mathcal{D}$, they minimize the risk over the \textit{adversarially perturbed} version of these examples. In essence, they augment the training data with adversarial examples. This process is called \textbf{adversarial training}, and it can be formalized using the following min-max objective:

            \begin{equation}\label{eq:saddle_point}
            \resizebox{.91\hsize}{!}{
            $\min\limits_{\theta} \rho(\theta), \quad \text{where} \quad \rho(\theta) = \mathds{E}_{(x,y) \sim \mathcal{D}} \Big[ \max\limits_{\delta: \hspace{0.2em} d(x+\delta,x)\leq\epsilon} \mathcal{L}(F_{\theta}(x + \delta), y) \Big]$}
            \end{equation}

            Note that the inner maximization is the adversary's objective from Equation~\ref{eqn:advrisk} and the outer minimization aims to make it harder for the adversary to achieve its objective. This can be viewed as the formalization of the defender's objective. In their work, Madry~\etal used the Projected Gradient Descent (PGD) attack, an iterative form of the FGSM attack~\cite{goodfellow2014explaining}, to find an approximate solution for the inner maximization objective. Their results showed that adversarial training creates MNIST and \cifar classifiers with high adversarial robustness compared to standard training.
            
            One major drawback of adversarial training is that it has a high computational cost. Standard training involves one forward and one backward pass through the classifier at every training iteration. Adversarial training requires several forward and backward passes per training iteration. For example, the model trained by Madry~\etal~\cite{madry2018towards} on \cifar requires $8$ forward and backward passes in total (7 for the inner maximization and 1 for the outer minimization). This slows down the training process significantly. Shafahi~\etal~\cite{shafahi2019adversarial} report that adversarially training a model on \cifar (similar to Madry~\etal) is $7\times$ slower than standard training.

\subsection{Transferring Adversarial Robustness} \label{sec:transferring}

    Ilyas~\etal~\cite{Ilyas2019AdversarialEA} discuss that adversarial examples are the result of \textit{non-robust features} that are born out of statistical patterns in the underlying data distribution. These features, although weakly correlated with the correct output, result in high predictive performance on non-adversarial images. Therefore, they can act as potential attack vectors for adversarial evasion attacks as this weak correlation can be easily manipulated using small amounts of perturbations in the image. \textit{Robust features}, however, have a strong correlation with the correct output and therefore are harder to manipulate under the given adversarial budget. Therefore, if one can encourage a model to learn robust features instead of non-robust ones, meaningful adversarial robustness can be achieved using standard training. To achieve this, Ilyas~\etal propose a method for removing non-robust features from images in a dataset. They begin with the assumption that models trained using adversarial training~\cite{madry2018towards} learn robust features. They then propose generating a \textit{robustified} version $x_r$ of any given image $x$ using the following optimization: 
    \begin{equation} \label{eqn:feature_loss}
        \min_{x_r} \quad ||g(x_r) - g(x)||_2
    \end{equation}
    where $g(\cdot)$ returns the penultimate layer outputs of an adversarially trained model. This optimization is solved using gradient descent with $x_r$ initialized using an image randomly sampled from the training data, independently of label of $x$. This ensures that $x_r$ has minimum amount of non-robust features correlated with the label of $x$ in expectation. Training a classifier using this robustified training data results in the model exhibiting significantly higher adversarial robustness, with a small loss in standard accuracy, as compared to a classifier trained on the original training data.

    The work of Ilyas~\etal~\cite{Ilyas2019AdversarialEA} is pivotal as it shows that non-trivial adversarial robustness can be achieved using the standard training framework. What they do, in essence, is \textit{transfer} an adversarially trained model's knowledge of robust features to another dataset. However, their robustification process still produces images with some amount of non-robust features as evidenced by empirical results. Models trained on the robustified data only exhibit meaningful adversarial robustness for small values of $\epsilon$. When evaluated against higher values of $\epsilon$, the model's performance on adversarial examples becomes worse than random chance. This phenomenon is further discussed in Section~\ref{sec:transfer}.

    \begin{figure*}[tb!]
        \centering
        \includegraphics[width=\textwidth]{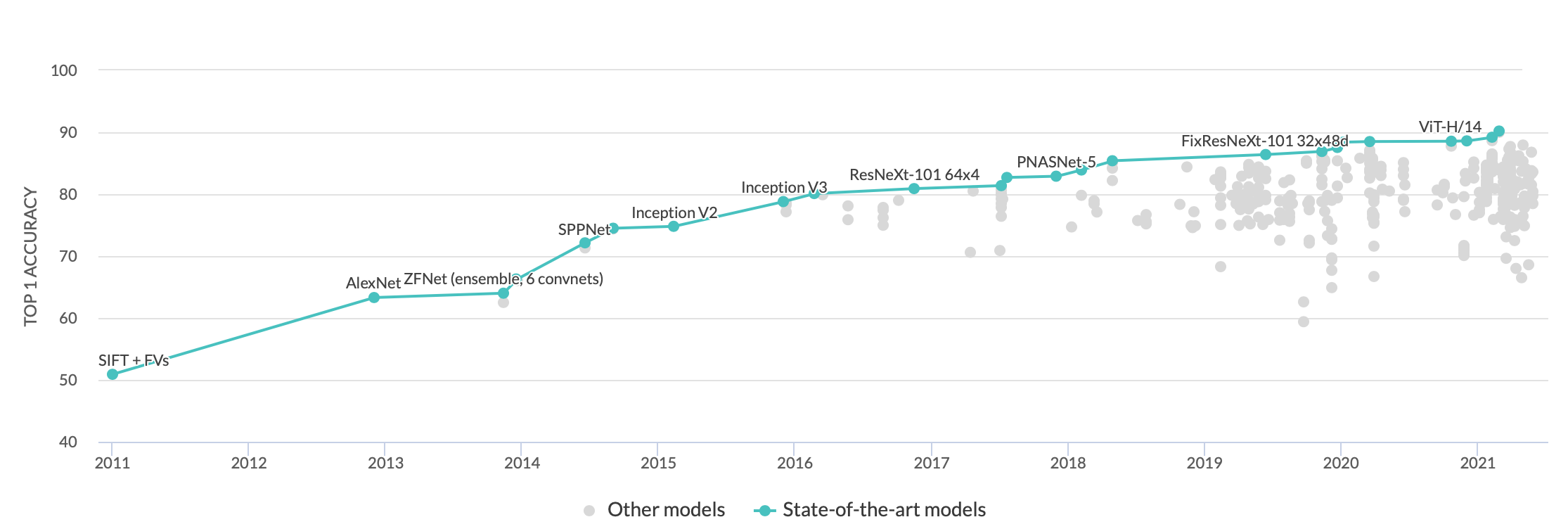}%
        \caption{State-of-the-art performance on \imagenet, a popular visual recognition benchmark, over the last decade~\cite{imagenet_benchmark}.}
        \label{fig:imagenet_benchmark}
    \end{figure*}

\section{Why Transfer Adversarial Robustness?}

    
    The rise of deep learning has been accompanied by its widespread adoption in commercial systems. Autonomous driving systems, news aggregators, virtual assistants, voice recognition, and fraud detection systems are just some of the systems we rely on every day that use deep learning models to accomplish their tasks. Even in their current imperfect state~\cite{marcus2018deep}, deep learning models are experiencing rapid improvements, and much of their potential is believed to be unrealized~\cite{forbes}. Like other innovations in computer systems, the increased adoption of deep learning puts a spotlight on their reliability and robustness against adversaries whose goal is to cause the system to misbehave. Early examples of these concerns have manifested as physically realizable adversarial attacks on deep learning models~\cite{eykholt2018robust}.  For safety-critical commercial systems using deep learning models (\eg self-driving cars), robustness against such attacks is highly desirable. We argue that to be usable in real world systems, the process of robustification of deep learning models should have the following characteristics:
    
    \begin{itemize}
        \setlength\itemsep{1em}
        \item \paratitle{Low standard accuracy reduction} Tsipras~\etal~\cite{tsipras2018robustness} established that there is a trade-off between a model's performance on adversarial and non-adversarial (\ie natural) inputs. In real world scenarios, adversarial attacks are expected to be rare anomalous occurrences, much like the instances of fraud in credit card transactions. Therefore, commercial systems with strict requirements for standard accuracy (\ie accuracy on natural inputs) will be hesitant to adopt solutions that significantly impact their functionality in the absence of an adversary. Taking this into consideration, a robustification process should minimally reduce standard accuracy while conferring non-trivial adversarial accuracy.
        
        \item \paratitle{Low amortized training cost} In production, the life-cycle of a deep learning model involves regular re-training and fine-tuning because of the availability of new data and improvements in the models. For example, Figure~\ref{fig:imagenet_benchmark} presents the rate at which the state-of-the-art accuracy on \imagenet dataset has improved in the last decade. The $~40\%$ gain in the accuracy during this period results from hundreds, if not thousands, of iterative improvements and modifications and our increased capacity to train deeper models. Adoption to a rapidly evolving environment mandates any robustification technique to impose minimal average overhead in terms of training cost and scale well with the complexity of the newer generation of models.
    \end{itemize}

    As one of the most recognized defense approaches, adversarial training, in its current formulation, is poorly suited for real world use. While the models trained using adversarial training show significant adversarial robustness~\footnote{The robustness of adversarial training is empirically validated against first-order adversaries.}, the training overhead remains an issue. Due to the need to generate adversarial examples on the fly during training, multiple forward and backward passes are made, making adversarial training computationally expensive. Furthermore, the training overhead only gets worse as the model and data increase in complexity and dimensionality. In our experiments, adversarially training a classifier (using $7$-step PGD attack) was on average $7\times$ slower than standard training. Any modifications in the model requires a complete repeat of this expensive process as reuse of the existing adversarially trained model is not possible.

    \sys seeks to improve the usability of adversarial training by allowing adversarial robustness to transfer across models, thereby eliminating the cost of repeated adversarial training. Through enabling transferable adversarial robustness, our proposed method allows for the reuse of older adversarially trained models to train new adversarially trained models using a modified ERM objective. Furthermore, we can exploit our approach to speed up standard adversarial training at the cost of a small amount of adversarial accuracy by first adversarially training a smaller, faster model and then transferring the robustness to a larger, slower model. 

\section{\syslong} \label{sec:method}
    Our objective is to make the process of training adversarially robust models computationally efficient. To this end, we propose \textit{\syslong} (\sys), a student-teacher framework to transfer adversarial robustness between models. \sys allows us to train adversarially robust models at a computational cost comparable to standard training. Empirically, we show that our method helps models attain high standard and adversarial accuracy in the smallest training time as compared to prior works~\cite{shafahi2019adversarial,Goldblum_2020}. 
    
    Our design begins with the same assumption as Ilyas~\etal~\cite{Ilyas2019AdversarialEA}, \ie advesarially trained models learn robust features. Thus, given an adversarially trained teacher model $T_\phi$, we train a student model $S_\theta$ to match the teacher's penultimate layer representations on natural (\ie non-adversarial) images. This is done using the following training objective:
    \vspace{-.5em}
    \begin{equation} \label{eqn:rrm_loss}
    \begin{split}
    \min_{\theta} \mathds{E}_{(x,y) \sim \mathcal{D}} [\lambda\cdot\mathcal{L}_{CE}(S_\theta(x),y) + \mathcal{L}_{R}(g_S(x), g_T(x))]
    \end{split}
    \end{equation}

    The functions $g_S(\cdot)$ and $g_T(\cdot)$ return the penultimate layer representations of classifiers $S_{\theta}$ and $T_{\phi}$ respectively. Our loss function includes two terms: (1)~the cross-entropy loss $\mathcal{L}_{CE}$; and (2)~the \textit{robust representation loss} $\mathcal{L}_R$, which computes the distance between the penultimate layer representations of $S_{\theta}$ and $T_{\phi}$. Updating $\theta$ to minimize $\mathcal{L}_R$, while keeping $\phi$ frozen, forces $S_{\theta}$ to learn penultimate layer representations that resemble those of adversarially robust $T_{\phi}$. Matching the penultimate layer representations in this way allows us to transfer adversarial robustness from $T_{\phi}$ to $S_{\theta}$. Algorithm~\ref{alg:pseudocode} provides the pseudo-code of the \sys training method.

    The coefficient $\lambda$ is used to appropriately weigh the contribution of $\mathcal{L}_{CE}$ towards the total loss. The higher the value of $\lambda$, the more $S_{\theta}$ biases towards maximizing standard accuracy with a smaller focus on the teacher's robust representations, which in turn lowers the student's adversarial robustness. This observation is consistent with the findings of prior works~\cite{Ilyas2019AdversarialEA,tsipras2018robustness}. Thus, if we select a small value for $\lambda$, we can instead bias $S_{\theta}$ to focus more on robust representations and improve its adversarial robustness through the supervision provided by $\mathcal{L}_R$. The value of $\lambda$ can be tuned based on the user's requirement\footnote{Note that it is mandatory to set $\lambda$ to a non-zero value. Otherwise, no loss gradient will be present to train the final layer of $S_{\theta}$ as $\mathcal{L}_R$ does not depend on the final layer.}.

    As we are training $S_{\theta}$ with natural images only, the computational cost of our approach is comparable to that of standard training and much lower than adversarial training~\cite{madry2018towards}. In addition to the backpropagation step present in standard training, \sys requires an additional forward pass through classifier $T_{\phi}$ to compute $\mathcal{L}_{R}$, which we later show is only a small amount of additional overhead. Although adversarially training $T_{\phi}$ is necessary, its cost is amortized as we can robustly train future student models, as might be required in large-scale commercial systems. 

    \begin{algorithm}[hbt!]
        \caption{{\small\syslong (\sys)}} \label{alg:pseudocode}
        \KwInput{Training data distribution $\mathcal{D}$, learning rate $\eta$, training iterations $\mathcal{T}$}
        \KwOutput{Robust student classifier $S_\theta$}
        $T_\phi\gets\textsc{AdversarialTraining}(\mathcal{D})$ \;
        $S_\theta\gets$ random initialization \;
        $g_T\gets$ $T_\phi$'s mapping from input to penultimate layer \;
        $g_S\gets$ $S_\theta$'s mapping from input to penultimate layer \;
        $i \gets 1$ \;
        \While{i $< \mathcal{T}$}
        {
            Sample input batch $\{ (x_1,y_1), (x_2,y_2), \cdots, (x_n,y_n)\}$ from $\mathcal{D}$\;
            $l_{r}\gets \frac{1}{n} \sum_{j=1}^n \mathcal{L}_R(g_S(x_j), g_T(x_j))$ \;
            $l_{ce}\gets \frac{1}{n} \sum_{j=1}^n \mathcal{L}_{CE}(S_\theta(x_j), y_j)$ \;
            $l_{total}\gets \lambda \cdot l_{ce} + l_{r}$ \;
            $\theta \gets \theta - \eta \cdot \nabla_{\theta} l_{total}$ \;
            $i \gets i + 1$ \;
        }
    \end{algorithm}
    
    \subsubsection*{Why match penultimate layer representations?}
    Goldblum~\etal~\cite{Goldblum_2020} demonstrated that it is possible to transfer adversarial robustness from an adversarially trained teacher to a student using the traditional Knowledge Distillation (KD) loss~\cite{hinton2015distilling}. The training objective of traditional KD is defined as follows:
    \begin{equation} \label{eqn:kd_loss}
        \min_{\theta} \mathds{E}_{(x,y) \sim \mathcal{D}} [ (1-\alpha)\mathcal{L}_{CE}(S_{\theta}^t(x), y) + \alpha t^2 \mathcal{L}_{KL}(S_{\theta}^t(x), T_{\phi}^t(x)) ]
    \end{equation}
    Here, $\mathcal{L}_{KL}$ is the KL divergence loss and is applied between the temperature (or $t$) scaled softmax outputs of the student and the teacher. The hyperparameter $\alpha$ is used to control the contribution of $\mathcal{L}_{KL}$ and the standard cross-entropy loss $\mathcal{L}_{CE}$ towards the total loss. This training objective is similar to the training objective of \sys (Equation~\ref{eqn:rrm_loss}) with the key difference being the layer whose outputs are being matched. While KD traditionally matches the final layer output of the student and the teacher, \sys instead matches the learned representations (\ie the penultimate layer output). We recognize that, as a result of minimizing the cross-entropy loss during training, a highly accurate teacher's outputs will closely resemble the ground truth labels. Therefore, matching the final layer outputs may limit the information transferred to the student as opposed to using the learned representations.
    In Section~\ref{sec:transfer}, we compare the KL divergence loss used on final layer outputs with our proposed robust representation loss and demonstrate the improved performance when the learned representations are used to transfer knowledge. We will further discuss the differences between \sys and the KD method used by Goldblum~\etal in Section~\ref{sec:related}.


    Matching the penultimate layer representations also allows us to preserve our \textbf{model agnostic} design. While the intermediate representations from an arbitrary layer may be higher in dimensionality, the representation matching approach used by \sys requires that the student and teacher's representations are the same dimensionality. Different model architectures can vary highly with respect to the dimensions of their intermediate layers, and thus, would require invasive changes at the intermediate layer for representation matching to be used. Furthermore, these changes would need to be cascaded downstream. In choosing the penultimate layer, we can ensure that \sys is model agnostic as most popular architectures (VGG, ResNet, \etc) have identical penultimate layer dimensions by default. In cases where the dimensions differ, we can simply add a single layer after the penultimate layer to one of the models to ensure proper sizing without the need for downstream changes. Our experiment results presented in Section~\ref{sec:results} demonstrate the effectiveness of this approach.

    \section{Threat Model}
        In this section, we provide specifications of the threat model under which we perform our evaluations. Our threat model is similar to the one used by Madry~\etal~\cite{madry2018towards}. Additionally, we claim similar adversarial robustness guarantees as them.
        
        \paratitle{Adversary Goals}
            In this work, we focus on evasion attacks on image classifiers. The adversary's goal is to imperceptibly perturb a given image such that the resulting image is misclassified by the image classifier. Evasion attacks are of two types: \textit{targeted} and \textit{untargeted}. Targeted evasion attacks seek to perturb images so that the classifier outputs a specific label that is desirable for the adversary. In untargeted evasion attacks, all incorrect labels are of similar value to the adversary. The objective of an untargeted attack is formalized in Equation~\ref{eqn:advrisk}. In this paper, we only evaluate against untargeted evasion attacks. However, based on the work by Madry~\etal~\cite{madry2018towards}, we can safely assume robustness against targeted evasion attacks as well.

        \paratitle{Adversarial Capabilities}
            The adversary is allowed to imperceptibly perturb the input to an image classifier. Similar to related works, we define the imperceptibility condition using $\ell_p$-norm, \ie $||\delta||_p \leq \epsilon$. The adversary uses the first-order gradients of the classifier to solve Equation~\ref{eqn:advrisk},  as the majority of optimization problems in machine learning are solved using first-order methods.

        \paratitle{Adversary Knowledge}
            We evaluate under the white-box threat model and assume that the adversary has complete knowledge of the classifier and its parameters. Additionally, the adversary is aware of the defense algorithm and can adapt to it. The adversary also has access to the training data used to train the target classifier.

\section{Evaluation}\label{sec:results}

    We conduct experiments to demonstrate the superiority of \sys along two dimensions: (1) training time (Section~\ref{sec:speedup}), and (2) effectiveness of transfer (Section~\ref{sec:transfer}). We perform both set of experiments using the \cifar dataset and compare against most relevant recent prior works. We also demonstrate that \sys scales to high dimensional datasets using the \rimagenet dataset~\cite{tsipras2018robustness} (Section~\ref{sec:rimagenet}). Through our evaluation, we verify that the results presented are statistically significant. For adversarial accuracy computation, we follow the standard practice in adversarial machine learning literature and use multiple random restarts to ensure that the attack doesn't get stuck in bad local maxima. For epoch timings, we compute the $95\%$ confidence interval to study statistical significance of our speedup results.

    \begin{table*}[t!]
        \centering
        \caption{Comparing the performance and training time of a robust \resnetl trained with different approaches. The teachers used for \sys models are noted in the parentheses. The adversarial accuracy evaluation is done using an $\ell_\infty$-bound AutoPGD attack~\cite{croce2020reliable} with $\epsilon=8/255$, 50 iterations and 10 random restarts. Compared to SAT, \sys achieves significant speedup while maintaining comparable adversarial accuracy and suffering minor drop in natural accuracy. Compared to Free AT, \sys achieves better natural and adversarial accuracy while converging $\sim 1.8\times$ faster. For epoch time, we report the 95\% confidence interval to demonstrate statistical significance.}
        \resizebox{2\columnwidth}{!}{%
        \begin{tabular}{@{}lcccccc@{}}
        \toprule
        \textbf{Method} & \textbf{Epochs} & \textbf{Epoch Time (sec)} & \textbf{Total Time (min)} & \textbf{Natural} & \textbf{AutoPGD} \\
        \midrule
        SAT & 150 & 723.03 $\pm$ 0.88 & 1807.58 & 85.50\% & 48.38\% \\
        Fast AT & 40 & 289.16 $\pm$ 0.22 & 192.77 & 83.73\% & 50.47\% \\
        \hline
        Free AT & 96 & 36.58 $\pm$ 0.23 & 58.44 & 77.74\% & 45.20\% \\
        Free AT & 48 & 36.58 $\pm$ 0.03 & 29.22 & 71.28\% & 41.53\% \\
        \hline
        \sys (\vggs) & 48 & 37.78 $\pm$ 0.09 & 30.22 & 76.17\% &  49.30\% \\
        \sys (\resnets) & 48 & 39.78 $\pm$ 0.10 & 31.82 & 80.32\% & 48.67\% \\
        \bottomrule
        \end{tabular}
        }
    \label{tab:cifar_res1}
    \end{table*}

    \subsection{Experimental Setup}
    All of our experiments were performed using the PyTorch library~\cite{NEURIPS2019_9015}.  Mixed-precision training was performed using the Nvidia Apex library~\cite{nvidiaapex}. We follow prior works for choosing the hyperparameters used in our experiments (details in Appendix~\ref{app:hyperparams}). We train a \resnetl model using different robustification approaches and compare the adversarial robustness of the resulting models to measure the relative effectiveness of these approaches.
    We measure adversarial robustness at test time using the AutoPGD attack~\cite{croce2020reliable}. The attack uses the cross-entropy loss and is run for 50 iterations with 10 random restarts (adopted from Wong~\etal~\cite{wong2019fast}). We use the IBM Adversarial Robustness Toolbox (ART)~\cite{art} to perform the attack. For \sys models, we use the \textbf{cosine similarity} metric to compute the robust representation loss $\mathcal{L}_R$ in Equation~\ref{eqn:rrm_loss}. The code supporting our experiments is available at \url{https://github.com/Ethos-lab/robust-representation-matching}. Our code is based on the implementation provided by Wong~\etal~\cite{wong2019fast}~\footnote{\url{https://github.com/locuslab/fast_adversarial}} and MadryLab's robustness package~\cite{robustness}~\footnote{\url{https://github.com/MadryLab/robustness}}.

    \paratitle{Hardware} We ran our experiments on two different machines. The \cifar experiments were run on a machine with an Intel Xenon(R) Gold 6136 CPU, 16 GB RAM, and an Nvidia Titan V GPU. The \rimagenet experiments were run on a second machine with an Intel Xenon(R) E5-2690 CPU, 16GB RAM, and an Nvidia V100 GPU. Due to GPU memory limitations on the second machine, the \vggm experiments were run on the first machine across 2 GPUs - Nvidia Titan V and GeForce RTX 2080 Ti.

    \subsection{Adversarial Training Speedup} \label{sec:speedup}

    In this section, we demonstrate how \sys can be used to speed up adversarial training and compare it against two adversarial training approaches: (1) Standard Adversarial Training (SAT), proposed by Madry~\etal~\cite{madry2018towards} and (2) Free Adversarial Training, recently proposed by Shafahi~\etal~\cite{shafahi2019adversarial} to speed up SAT. 
    We apply the DAWNBench improvements\footnote{Mixed-precision training with cyclic learning rate scheduling.} proposed by Wong~\etal~\cite{wong2019fast} to both \sys and Free AT during the experiments and also show its effects on standard adversarial training (Fast AT).
    Overall we demonstrate that, given a pre-trained teacher, \textbf{\sys achieves adversarial robustness comparable to SAT in the least amount of training time}. 

    We compare the performance and the time required to train an adversarially robust \resnetl model with each approach. Using the attack budget from prior work, we conduct the experiments using an $\ell_\infty$-bound adversary with $\epsilon=8/255$. For \sys, we use $\lambda=5e-3$ and provide results using \vggs and \resnets as teachers to demonstrate \sys's capability in transferring robustness across different class of model architectures. Additionally, to demonstrate that \sys is model-agnostic, we purposely use student-teacher pairs with different penultimate layer dimensions. Following the strategy discussed in Section~\ref{sec:method}, we remedy the dimensional mismatch by adding a single fully connected layer to the model with higher penultimate layer dimension. For details regarding classifier modifications we make, see Appendix~\ref{app:pen_dim}. The teachers are trained using Fast AT to reduce the teacher training overhead. With respect to the number of training epochs, we report the performance of SAT and \sys at convergence and report the performance of Fast AT and Free AT with their default parameters. We also include the performance of Free AT at the same number of epoch as \sys convergence ($48$ epochs) for side-by-side comparison. The summary of results is presented in Table~\ref{tab:cifar_res1}. 


    \subsubsection{Standard Adversarial Training (SAT)}
    As we see in Table \ref{tab:cifar_res1}, \sys attains adversarial robustness comparable to SAT and Fast AT but in a fraction of the time. Specifically, \sys achieves an average speedup of $\sim58\times$ over SAT and of $\sim6\times$ over Fast AT. We note that the performance of \sys models on natural images is reduced by $\sim 8\%$ when using a \vggs teacher and $\sim 4\%$ when using a \resnets teacher. In Section~\ref{sec:discussion}, we discuss the balance between natural and adversarial accuracy based on the value of $\lambda$ used during training.

    

    \subsubsection{Free Adversarial Training (Free AT)}
    Free AT speeds up SAT by requiring only a single forward and backward pass during each training iteration~\cite{shafahi2019adversarial}. A single backward pass is used to compute the gradients of the loss with respect to the model's parameters (to train the model) and the input image (to compute the adversarial perturbation). The PGD attack used in SAT requires several backward passes to compute the adversarial perturbation. Free AT mimics this by repeating the same batch $m$ times and using the adversarial perturbation computed in one iteration to initialize the adversarial perturbation of the next iteration. After $m$ replays, a new batch is used and the adversarial perturbation is reset.

    For brevity, we only compare against the version of Free AT~\cite{shafahi2019adversarial} with DAWNBench improvements applied as it is comparable to Free AT in terms of adversarial robustness, but superior in terms of total training time. We use $m=8$ as proposed by the authors. Compared to \sys, Free AT is slightly faster with respect to per epoch time as shown in Table \ref{tab:cifar_res1}. However, \textbf{\sys achieves better performance} ($\sim 2.5\%$ higher standard accuracy and $\sim 3.5\%$ higher adversarial accuracy) and \textbf{has significantly lower total training time} as it converges $\sim 1.8\times$ faster than Free AT.

    \begin{figure}[b!]
        \centering
        \includegraphics[width=\columnwidth]{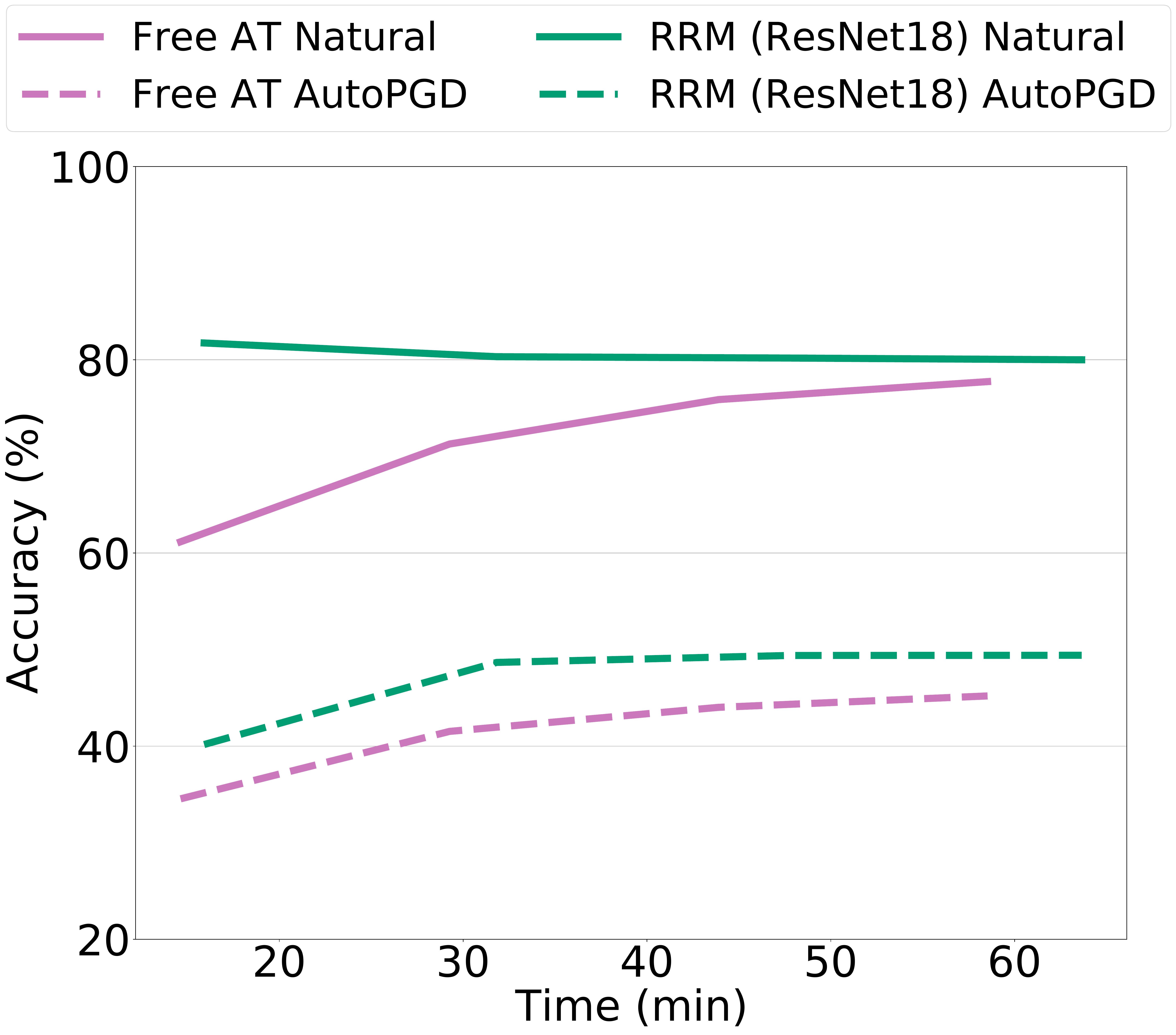}%
        \caption{Plotting the performance of a \resnetl model trained using Free AT and \sys for different amount of training time budget. The $x$-axis represents the total time (in minutes) the model was trained for and the $y$-axis represents the accuracy of the trained model. Each data-point in the curve is the average model performance across 3 independent training runs. \sys outperforms Free AT while converging faster.}
        \label{fig:cifar_epoch_vs_acc}
    \end{figure}

    In Figure \ref{fig:cifar_epoch_vs_acc}, we plot the performance of a \resnetl model trained using Free AT and RRM for different number of epochs. The $x$-axis represents the total number of epochs the models were trained for and the $y$-axis represents the corresponding accuracy. Solid lines represent accuracy on natural test set and dashed lines represent accuracy on adversarial test set generated using AutoPGD attack with 50 steps and 10 random restarts. Each data-point in the curve is the average model performance across 3 independent training runs. As can be seen, \sys converges faster and has better performance across the entire range of the $x$-axis. 

    \subsubsection{Teacher Overhead in \sys}
\begin{figure}[h!] 
        \centering
        \includegraphics[width=\columnwidth]{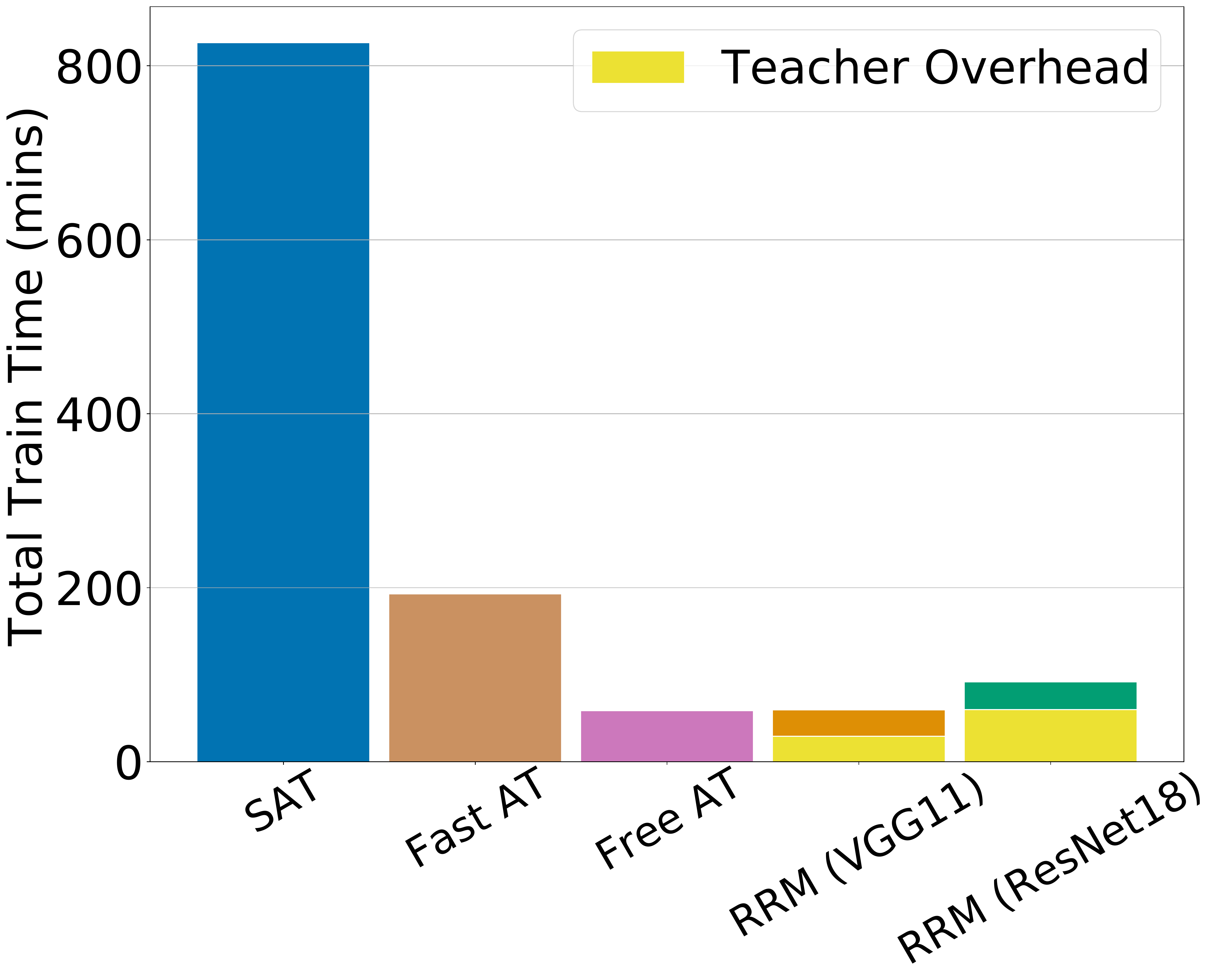}%
        \caption{Comparing total training times of SAT, Fast AT, and Free AT with \sys. Yellow regions represent the total time of adversarially training a teacher. If an adversarially robust teacher is already trained, the total training time of \sys is decreased significantly.}
        \label{fig:cifar_speed}
    \end{figure}

    The results reported in Table~\ref{tab:cifar_res1} only present the time required to train the \resnetl model. In case of \sys, we do not include the time required to train the teacher. We argue that the cost to train the teacher can largely be amortized as one teacher training session can be leveraged to train an arbitrary number of students. In Figure~\ref{fig:cifar_speed}, however, we compare the training time of \sys with other methods when the teacher's training overhead is included (for numerical results see Appendix~\ref{app:speedup_results}). In this setting, \sys is still on average $17.5\times$ and $2.7\times$ faster than SAT and Fast AT, respectively. When compared to Free AT, \sys requires comparable total training time to train a model with better performance.

    \subsection{Adversarial Robustness Transfer} \label{sec:transfer}
    
    In this section, we compare against two prior works that propose techniques to transfer adversarial robustness between models. First, we examine the robust data training approach~\cite{Ilyas2019AdversarialEA}, which creates a robust dataset learned from an adversarially trained model to transfer adversarial robustness. Robustness is transferred to other models through standard training on the robust dataset. Second, we reproduce the experiment conducted by Goldblum~\etal~\cite{Goldblum_2020} that suggested an adversarially trained teacher's robustness can be transferred to a student using knowledge distillation. Given the pre-trained teacher, they included a new loss term, a KL divergence loss between the temperature scaled softmax outputs of the teacher and student models, during standard training. Our empirical results demonstrate that \textbf{\sys is superior to both these approaches with respect to the effectiveness of the transfer}.
    
    We compare the performance of an adversarially robust \resnetl model. To remain consistent with the work by Ilyas~\etal~\cite{Ilyas2019AdversarialEA} and lessen hyperparameter tuning, we use an $\ell_2$-bound adversary with $\epsilon=1.0$. For \sys, we use $\lambda=5e-5$. The results are summarized in Table~\ref{tab:cifar_res2}.
    
    \begin{table}[t!]
        \centering
        \caption{Comparing \sys against RDT~\cite{Ilyas2019AdversarialEA} and KD~\cite{Goldblum_2020} using \resnetl. Performance of model trained using SAT is provided for reference. The adversary is $\ell_2$-bound with $\epsilon=1.0$. The evaluation is done using AutoPGD attack~\cite{croce2020reliable} with 50 iterations and 10 random restarts. Models trained using \sys exhibit performance comparable to SAT and significantly better than model's trained using RDT and KD.}
        \resizebox{.9\columnwidth}{!}{%
        \begin{tabular}{@{}cccc@{}}
        \toprule
        \textbf{Method} & \textbf{Teacher} & \textbf{Natural} & \textbf{AutoPGD} \\
        \midrule
        SAT & - & 82.97 & 48.49 \\
        \hline
        \multirow{2}{*}{RDT} & \vggs & 74.61 & 1.10 \\
          & \resnets & 80.47 & 1.22 \\
        \hline
        \multirow{2}{*}{KD} & \vggs & 80.12 & 20.89 \\
          & \resnets & 83.72 & 2.83 \\
        \hline
        \multirow{2}{*}{\sys} & \vggs & 78.53 & 47.24 \\
          & \resnets & 80.80 & 46.18 \\
        \bottomrule
        \end{tabular}
        }
    \label{tab:cifar_res2}
    \end{table}

    \subsubsection{Robust Data Training (RDT)}
     Ilyas~\etal~\cite{Ilyas2019AdversarialEA} propose \textbf{dataset robustification} in order to transfer robustness between models. Their method removes the \textit{non-robust features} from the training dataset through an optimization process (Section~\ref{sec:transferring}) resulting in a new ``robustified'' dataset. They demonstrated that the ``robustified'' training data can be used  with standard training to train classifiers with non-trivial adversarial robustness. While both their work and ours use an adversarially robust classifier's penultimate layer to identify robust features, their work adds an additional intermediate dataset robustification step. On the other hand, our work adds a feature loss to the standard training loss to directly encode the robust features into the student.

    To generate a robust dataset from an adversarially trained classifier, we follow the steps described by Ilyas~\etal. First, we use a random image from the training set to initialize optimization. Then, $1000$ steps of gradient descent with a step size of $0.1$ are performed to minimize the loss described in Equation~\ref{eqn:feature_loss}. At each step, the $\ell_2$-norm of the gradient is normalized. We used an $\ell_2$-bound adversary with $\epsilon=1.0$ to adversarially train the classifier used for robustification (\ie teacher model).

    In Table~\ref{tab:cifar_res2}, we observe that models trained using \sys have comparable standard accuracy and significantly better adversarial accuracy than models trained using RDT. In the original paper, RDT trained models exhibited non-trivial adversarial robustness with respect to smaller values $\epsilon$ of epsilon, which is what we originally observed as well. When we re-evaluated the models with respect to $\epsilon=1.0$ (\ie the value used to train the teacher), the student models exhibit negligible adversarial robustness. Furthermore, we found that the robust data generation process was computationally expensive. Using the proposed hyper parameters and an adversarially trained \resnets model, the robust data generation process took approximately $6$ hours to complete. Although this cost would be amortized, the poor performance of the student models suggest that this approach is not feasible for transferring robustness.

    \subsubsection{Knowledge Distillation (KD)}
    Prior to our work, Goldblum~\etal~\cite{Goldblum_2020} demonstrated that adversarial robustness can be transferred from an adversarially trained teacher to a student using knowledge distillation~\cite{hinton2015distilling} with naturally trained images. KD seeks to minimize the KL divergence between the temperature scaled softmax outputs of a student and a teacher model in addition to minimizing the student's classification loss (see Equation~\ref{eqn:kd_loss}). We reproduce this experiment based on information provided by Goldblum~\etal in their paper. We train both the student and the teacher with temperature $t=30$, the proposed value. The standard and adversarial accuracy (against a 20 steps PGD attack) of the adversarially trained teacher models are: (1) 78.3\% and 47.67\% for \vggs; and (2) 82.38\% and 51.15\% for \resnets. The performance of the student models is provided in Table~\ref{tab:cifar_res2}. Note that while training the student, we set the value of $\alpha$ to 1 as this corresponds to the \textbf{maximum} attainable adversarial robustness using this method. 

    When we adversarially attacked the student model with $t=1$ (\ie the default used during evaluation), we observed a non-trivial adversarial robustness similar to what Goldblum~\etal originally reported. However, when we set $t=30$ (\ie the value used during training), the student's adversarial accuracy drops significantly. This phenomenon is due to vanishing gradients originally observed by Carlini and Wagner~\cite{carlini2016defensive} when analyzing another distillation based defense~\cite{papernot2016distillation}. Thus, traditional KD is not feasible for transferring robustness.

        \begin{table*}[t!]
            \centering
            \caption{Comparing the performance and training time of a robust \resnetl and \vggm models trained using SAT and \sys. An \alexnet model trained using SAT is used as teacher for \sys. The adversarial accuracy evaluation is done using an $\ell_2$-bound AutoPGD attack~\cite{croce2020reliable} with $\epsilon=3$, 20 iterations, and 5 random restarts.}
            \resizebox{1.8\columnwidth}{!}{%
            \begin{tabular}{@{}lcccccc@{}}
            \toprule
            \textbf{Method} & \textbf{Epochs} & \textbf{Epoch Time (mins)} & \textbf{Total Time (hrs)*} & \textbf{Natural} & \textbf{AutoPGD} \\
            \midrule
            \multicolumn{6}{c}{\textbf{\resnetl}} \\
            \midrule
            SAT & 150 & 101.49 & 253.71 & 95.47\% & 84.36\% \\
            \sys & 60 & 13.66 & 13.66 & 88.19\% & 67.90\% \\
            \midrule
            \multicolumn{6}{c}{\textbf{\vggm}} \\
            \midrule
            SAT & 150 & 160.01 & 400.01 & 92.40\% & 80.91\% \\
            \sys & 60 & 33.30 & 33.30 & 87.86\% & 73.30\% \\
            \bottomrule
            \end{tabular}
            }
            \caption*{\textit{*The \resnetl and \vggm models were trained on different machines due to memory constraints.}}
        \label{tab:imgnet_res}
        \end{table*}

    \subsection{Scaling \sys to Complex Datasets} \label{sec:rimagenet}

        To examine how well \sys adapts to more complex datasets, we evaluate \sys using the \rimagenet dataset, which was introduced by Tsipras~\etal~\cite{tsipras2018robustness} to facilitate adversarial robustness research with high resolution images. The large number of classes present in the original \imagenet dataset makes it difficult to use SAT and achieve acceptable performance on natural and adversarial images. \rimagenet is generated by grouping together a subset of semantically similar classes from \imagenet into 9 super-classes. For our experiments we follow Ilyas~\etal~\cite{Ilyas2019AdversarialEA} and use an $\ell_2$-bound adversary with $\epsilon=3.0$. We train a \resnetl and a \vggm model using SAT and \sys and compare their performance and training times. For \sys, we use $\ell_2$ loss to compute robust representation loss $\mathcal{L}_R$, $\lambda=1e-3$, and an \alexnet teacher trained using SAT.


        In Table~\ref{tab:imgnet_res}, we compare the standard and adversarial accuracy of the \sys models against their SAT baselines. The \sys models exhibit competitive natural and adversarial accuracy, coming within a few percentage points of the corresponding SAT model's performance. Specifically, across the two models, there is an average reduction of $\sim 6\%$ in natural accuracy and of $\sim 12\%$ in adversarial accuracy. However, relative to SAT, \sys achieves a speedup of $\sim 18\times$ on \resnetl and $\sim 12\times$ on \vggm. All models were trained till convergence. When including the teacher's training time, \sys achieves a speedup of $5.4\times$ on \resnetl and $6.0\times$ on \vggm (for numerical results see Appendix~\ref{app:speedup_results}). Note that we do not use DAWNBench improvements in this set of experiments. The standard adversarial training performance of all models used in Table~\ref{tab:imgnet_res} is reported in in Appendix~\ref{app:sat_results}.


\section{Discussion}
\label{sec:discussion}
    In this section we discuss some noteworthy points regarding \sys. First, we discuss how changing the value of $\lambda$ affects the performance of the student model, which can inform users of \sys how to tune $\lambda$ based on their use case. Second, we explore the decreasing effectiveness of \sys  when the teacher is more complex than the student. In our presented results, the teacher's architecture was always less complex (\ie faster to train) than the student. Using a \resnetl teacher, we train multiple student models of decreasing complexity and observe lowering rates of robustness transfer.

    \subsection{Tuning the $\lambda$ Parameter}
\begin{figure}[t!]
        \centering
        \includegraphics[width=\columnwidth]{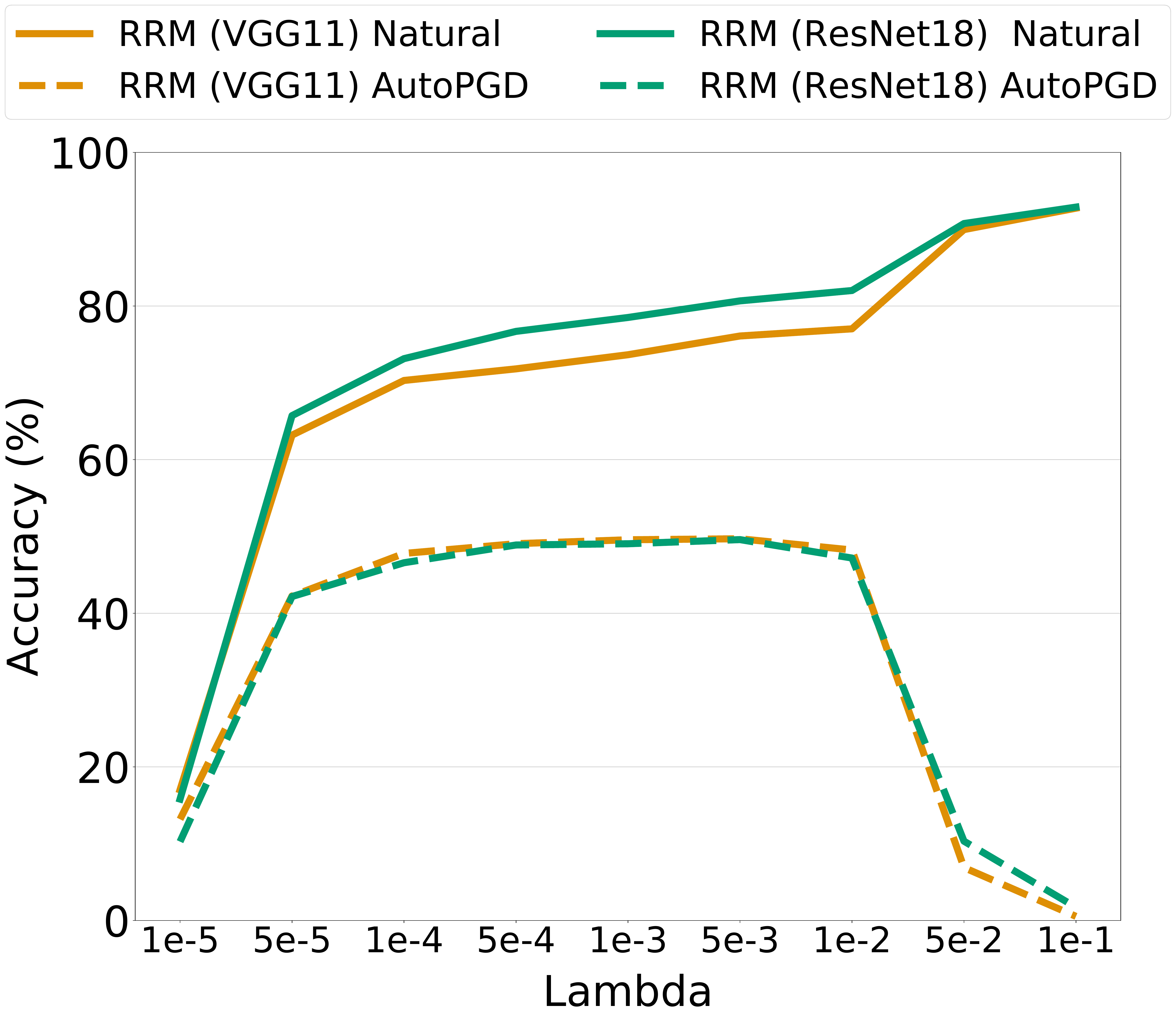}%
        \caption{Plotting the performance of \resnetl trained with \sys using two different teachers while varying the value of $\lambda$. We see that there is a plateau in adversarial accuracy for $5e-5\leq\lambda\leq1e-2$.  $\lambda$ values outside of this range either result in a model with poor natural accuracy and/or poor adversarial robustness.}
        \label{fig:cifar_lambda_tradeoff}
    \end{figure}

    In Figure~\ref{fig:cifar_lambda_tradeoff} we plot the performance of two \resnetl models trained using \sys, while varying the value of $\lambda$ used. The two models are trained with a \vggs and a \resnets teacher, respectively. The adversarial accuracy is computed using AutoPGD attack with 20 iterations and 5 random restarts.
    
    Equation \ref{eqn:rrm_loss} contains two loss terms: 1) $\mathcal{L}_{CE}$, which improves the model's natural accuracy and 2) $\mathcal{L}_R$, which encourages the model to learn the teacher's robust representations. As we decrease the value of $\lambda$, the contribution of $\mathcal{L}_{CE}$ is reduced, which increases the contribution of $\mathcal{L}_R$. In Figure~\ref{fig:cifar_lambda_tradeoff}, we observe this effect for $\lambda\geq1e-2$. For $5e-5\leq\lambda\leq1e-2$, we observe somewhat of a plateau in performance for the adversarial accuracy, with only a slight negative slope in natural accuracy. Thus, any value in this range will likely result in robust student model. Finally, if $\lambda$ becomes too small (\ie $\lambda< 5e-5$), the training focuses too much on matching the robust representation of the teacher that the student model's performance plummets. The drop in adversarial accuracy at this point is attributed to the model's poor natural accuracy rather than a decrease in robustness. 
    

    \subsection{Limit Testing}
\begin{figure}[t!]
        \centering
        \includegraphics[width=\columnwidth]{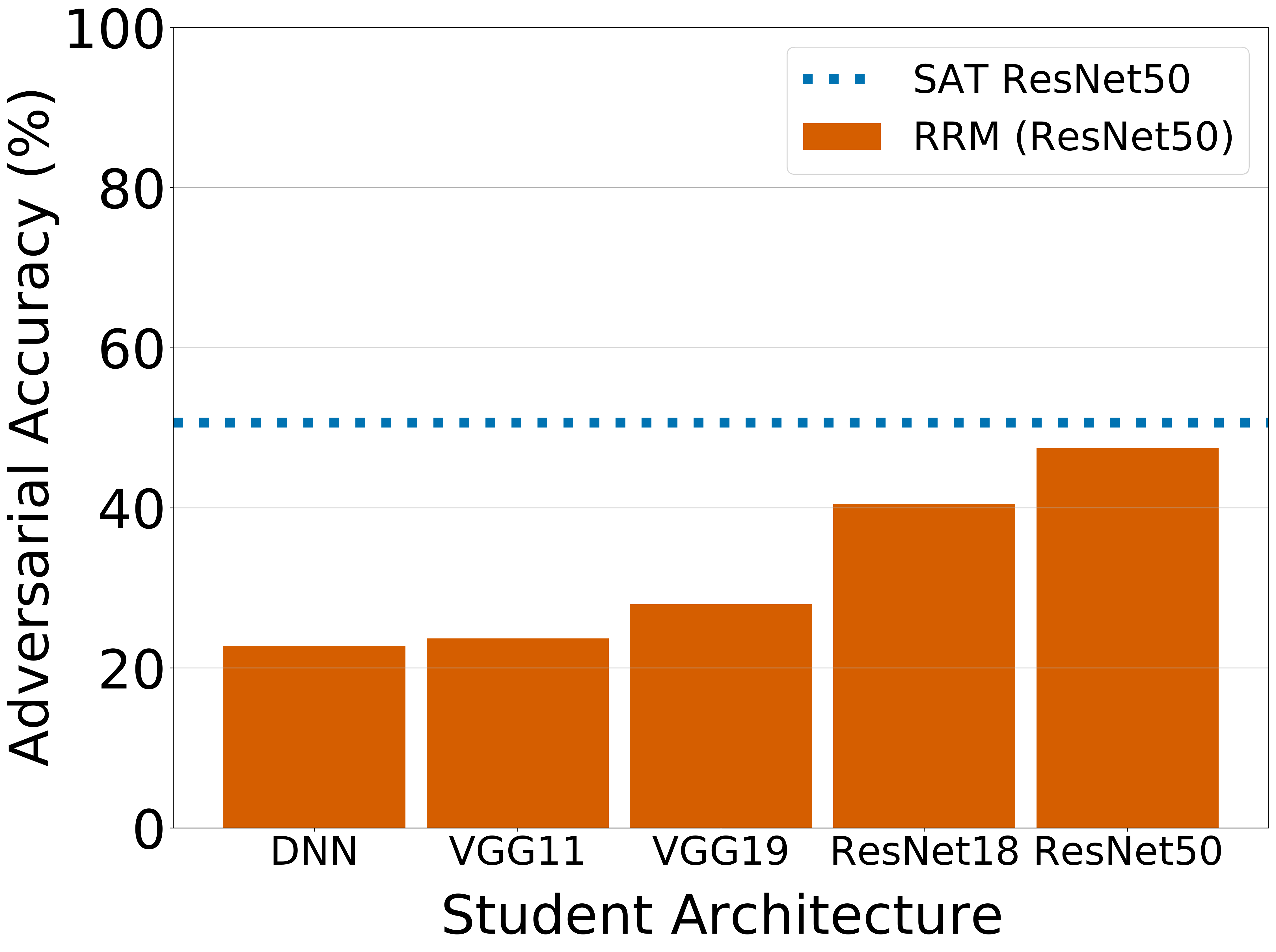}%
        \caption{Limit testing \sys on \cifar by transferring adversarial robustness from a \resnetl teacher to several students of varying complexities. Adversarial images were generated using an $\ell_2$-bound AutoPGD attack~\cite{croce2020reliable} with $\epsilon=1.0$, iterations $=20$, and 5 random restarts.}
        \label{fig:cifar_lt}
    \end{figure}

    Previously, we showed that \sys allows us to efficiently transfer a significant amount of robustness from an adversarially trained teacher to a student despite differences in architecture. However, during our experiments, we noticed a few cases of poor robustness transferability. For example, on \cifar, with a \resnetl teacher and an $\ell_2$-bounded adversary with $\epsilon=1$, the \vggs and \vggl students were only able to achieve $23.74\%$ and $28.03\%$ adversarial accuracy, respectively. Compare this with the \resnets and \resnetl student models, which achieve $40.56\%$ and $47.54\%$ adversarial accuracy. We observed a similar phenomenon with our \rimagenet experiments, in which an \alexnet model was only able to achieve $51.72\%$ adversarial accuracy (natural accuracy is $78.11\%$) when trained with a \resnetl teacher. In contrast, an \alexnet model trained with SAT achieved $75.24\%$ adversarial accuracy. These observations suggest that, under certain conditions, the effectiveness of \sys is limited. Figure~\ref{fig:cifar_lt} presents our exploration of these limits. We evaluated the adversarial accuracy on \cifar of several \sys models trained using the \resnetl teacher and compare them to the adversarial accuracy of the teacher. In addition to the model architectures we already mentioned, we also trained a simple DNN with two convolution layers and two fully connected layers. Using SAT, this DNN achieves $59.55\%$ natural accuracy and $34.15\%$ adversarial accuracy.

    If we rank the complexity of a classifier based on its per-epoch training time, then we have the following order: \resnetl > \resnets > \vggl > \vggs > DNN. Thus, we rank the \resnetl ($723$ seconds per-epoch) as the most complex and the DNN ($14.3$ seconds per-epoch) as the least complex classifier. Our results suggest that the simpler a student is compared to the teacher, the poorer the student's performance will be. We hypothesize that the per-epoch training time is a rough approximation of the model's expressive power. Thus, the robust features used by a complex teacher are harder for less complex students to learn, resulting in poor transferability in such cases. Further exploration is required to establish a concrete metric to predict the transferability between different classifier architectures when training with \sys. While these results suggest the existence of some limitations with our proposed approach, we note that \sys is still applicable in real world settings where there are relatively small differences in complexity between successive generations of model architectures. Furthermore, in cases where the student and teacher are trained from scratch, it is beneficial to pair a less complex teacher with a more complex student to reduce training overhead. The standard adversarial training performance of all models in Figure~\ref{fig:cifar_lt} are reported in Appendix~\ref{app:sat_results}.

\section{Related Works} \label{sec:related}
    \paratitle{Adversarially Robust Distillation (ARD)~\cite{Goldblum_2020}}
    In their work, Goldblum \etal used traditional knowledge distillation with an adversarially robust teacher to train an adversarially robust student. Here we describe the key differences between the traditional KD method from their paper and \sys. Note that we do not compare against ARD, which is the main contribution of their paper. This is because ARD requires training both the student and the teacher using adversarial training and solely focuses on improving adversarial robustness of the student without any regards to the total training time.
    
    First, RRM uses a cosine similarity loss focused on the models' penultimate layer (i.e., pre-logit layer) whereas Goldblum~\etal use a knowledge distillation loss focused on the models’ temperature scaled softmax outputs. Our approach encourages the student to utilize the robust representations learned by the teacher. When they have differently shaped representation layers, we add an additional layer of the correct shape after the current penultimate layer to one of the models. In contrast, KD expects the student to learn its own representations to match the teacher’s softmax output and requires an additional hyperparameter, the temperature t, as compared to \sys. Depending on its magnitude, $t$ can affect the success rate of gradient based adversarial attacks due to an artificial scaling of the logits~\cite{carlini2016defensive}. Note that the hyperparameter $\lambda$ in Equation~\ref{eqn:rrm_loss} and $\alpha$ in Equation~\ref{eqn:kd_loss} serve identical purpose of controlling the trade-off between natural and adversarial accuracy.
    
    The second difference is in intent. KD is traditionally used to transfer the performance of a larger more complex model to a smaller less complex model in an effort to ``compress'' the larger model into the smaller one. Goldblum \etal use this to \textit{improve the adversarial robustness} of smaller models by leveraging robust larger models and their experiments reflect this. Their approach augments adversarial training with KD to achieve high performance. As we showed in Section~\ref{sec:transfer}, distillation alone is insufficient to transfer the robustness of the teacher and, likely, only helps to fine-tune adversarial performance. In contrast, \sys seeks to \textit{reduce the adversarial training overhead} by leveraging smaller models to quickly train robust larger models. We necessarily do not adversarially train the student.

    
    \paratitle{Dataset Robustification~\cite{Ilyas2019AdversarialEA}} In their work, Ilyas \etal designed a dataset transformation approach to create a new dataset composed only of ``robust features''. With the robust dataset, one could use standard ERM (Equation~\ref{eqn:stderm}) to train a model with non-trivial adversarial robustness. The key difference between their approach and \sys lies in the technique used to transfer the robust features to the student model. Their work assumes that a model trained on a robustified dataset would automatically learn the robust features. As we showed in Section \ref{sec:transfer}, their method only marginally improves the adversarial robustness of the trained models. \sys directly provides the robust feature representations from the adversarially trained teacher to the student, which results in models with much higher adversarial accuracy. Furthermore, our method does not include the additional overhead of generating the robust dataset.

    \paratitle{IGAM~\cite{chan2020thinks}} Another work that explores transferring adversarial robustness was published by Chan~\etal~\cite{chan2020thinks}. Their approach for transferring adversarial robustness involves matching the student's loss gradients (with respect to the input) to the loss gradients of an adversarially robust teacher. While both their work and ours pertain to transferring adversarial robustness between models, their focus is on transferring adversarial robustness across task domains, which is orthogonal to the problem solved by \sys. They do not conduct experiments to transfer performance within the same task between models of different architectures. Instead, they focus on transferring robustness between models having same architecture, but trained for different tasks (using different datasets).

\section{Limitations and Future Work}
    \label{sec:future}
    
    \paratitle{\sys dependence on adversarial training}
    \sys requires an adversarially trained teacher model. Thus, any shortcomings of adversarial training, such as the large training overhead and potential overfitting~\cite{tramer2018ensembletraining} still exist with respect to the teacher model. However, we demonstrated that \sys reduces training time when an adversarially trained teacher is already available and in cases when training the student-teacher pair is significantly faster than adversarially training the student. Furthermore, any improvements to adversarial training  will improve \sys indirectly. In Section~\ref{sec:speedup}, we demonstrated that some of the speedup techniques proposed by Wong \etal~\cite{wong2019fast}, such as mixed-precision training, are compatible with \sys.

    \paratitle{\sys in other domains and model types} In this work, we only studied transferring adversarial robustness between deep neural networks in the image classification domain. It is unknown if \sys would work in other domains or with other types of models (e.g. decision trees or LSTMs). With respect to other domains, if adversarial training exists in the domain, we expect \sys to work as its core idea is to encourage the student to use the robust representations learned by a teacher. Regarding transferring between different model types, further investigation is needed to determine if this is feasible and if not, what modifications are required to make it feasible.

\section{Conclusion}
    Adversarial machine learning looms as an ever-present threat to the security and reliability of machine learning systems as research has proven attackers with a certain level of access can reliably cause them to misbehave. As such, it is desirable to train machine learning models that are robust to adversarial attacks in advance rather than wait for a breach to occur. Unfortunately, adversarial training, one of the most well-known and reliable defenses, is impractical to deploy in real world systems. Like software, machine learning models need to be constantly updated to maintain state-of-the-art performance due to the availability of new training data or the development of new model architectures. The high training overhead and poor scalability of adversarial training discourage users from adopting it as part of their training process.
    
    In this paper, we proposed a method to transfer adversarial robustness between models despite differences in their architectures. \sys enables low-cost, efficient transfer of robust representations learned by an adversarially trained teacher model to a new student model. By adding a new loss term to the standard training objective, an adversarially robust model can be trained on natural images only using \sys. On \cifar, we demonstrated that \sys outperforms state-of-the-art adversarial training speedup techniques. On \rimagenet, a higher dimensional dataset, we demonstrated that \sys remains effective both in terms of model performance and training speedup. 
    

\section*{Acknowledgement}
We thank Veena Krish, Farhan Ahmed, the anonymous reviewers, and our shepherd, David Freeman, for their valuable feedback. This work was supported by the Office of Naval Research under grants N00014-20-1-2858 and N00014-22-1-2001, and Air Force Research Lab under grant FA9550-22-1-0029. Any opinions, findings, or conclusions expressed in this material are those of the authors and do not necessarily reflect the views of the sponsors.

\bibliographystyle{plain}
\bibliography{main}

\begin{thebibliography}{10}

\bibitem{athalye2018obfuscated}
Anish Athalye, Nicholas Carlini, and David Wagner.
\newblock Obfuscated gradients give a false sense of security: Circumventing
  defenses to adversarial examples.
\newblock In {\em International Conference on Machine Learning (ICML)}, 2018.

\bibitem{carlini2016defensive}
Nicholas Carlini and David Wagner.
\newblock Defensive distillation is not robust to adversarial examples.
\newblock 2016.
\newblock arXiv:1607.04311.

\bibitem{carlini2017evaluating}
Nicholas Carlini and David Wagner.
\newblock Towards evaluating the robustness of neural networks.
\newblock In {\em IEEE symposium on security and privacy (S\&P)}, 2017.

\bibitem{chakraborty2018adversarial}
Anirban Chakraborty, Manaar Alam, Vishal Dey, Anupam Chattopadhyay, and Debdeep
  Mukhopadhyay.
\newblock Adversarial attacks and defences: A survey.
\newblock 2018.
\newblock arXiv:1810.00069.

\bibitem{chan2020thinks}
Alvin Chan, Yi~Tay, and Yew-Soon Ong.
\newblock What it thinks is important is important: Robustness transfers
  through input gradients.
\newblock In {\em Conference on Computer Vision and Pattern Recognition
  (CVPR)}, 2020.

\bibitem{cohen2019certified}
Jeremy~M Cohen, Elan Rosenfeld, and J.~Zico Kolter.
\newblock Certified adversarial robustness via randomized smoothing.
\newblock In {\em International Conference on Machine Learning (ICML)}, 2019.

\bibitem{croce2020reliable}
Francesco Croce and Matthias Hein.
\newblock Reliable evaluation of adversarial robustness with an ensemble of
  diverse parameter-free attacks.
\newblock In {\em International Conference on Machine Learning (ICML)}, 2020.

\bibitem{robustness}
Logan Engstrom, Andrew Ilyas, Hadi Salman, Shibani Santurkar, and Dimitris
  Tsipras.
\newblock Robustness (python library), 2019.
\newblock https://github.com/MadryLab/robustness.

\bibitem{eykholt2018robust}
Kevin Eykholt, Ivan Evtimov, Earlence Fernandes, Bo~Li, Amir Rahmati, Chaowei
  Xiao, Atul Prakash, Tadayoshi Kohno, and Dawn Song.
\newblock Robust physical-world attacks on deep learning visual classification.
\newblock In {\em International Conference on Learning Representations (ICLR)},
  2018.

\bibitem{Goldblum_2020}
Micah Goldblum, Liam Fowl, Soheil Feizi, and Tom Goldstein.
\newblock Adversarially robust distillation.
\newblock In {\em AAAI Conference on Artificial Intelligence}, 2020.

\bibitem{goodfellow2014explaining}
Ian~J Goodfellow, Jonathon Shlens, and Christian Szegedy.
\newblock Explaining and harnessing adversarial examples.
\newblock In {\em International Conference on Learning Representations (ICLR)},
  2014.

\bibitem{hinton2015distilling}
Geoffrey Hinton, Oriol Vinyals, and Jeff Dean.
\newblock Distilling the knowledge in a neural network, 2015.
\newblock arXiv:1503.02531.

\bibitem{art}
IBM.
\newblock {A}dversarial {R}obustness {T}oolbox ({ART}).
\newblock https://github.com/Trusted-AI/adversarial-robustness-toolbox.

\bibitem{Ilyas2019AdversarialEA}
Andrew Ilyas, Shibani Santurkar, Dimitris Tsipras, Logan Engstrom, Brandon
  Tran, and Aleksander Madry.
\newblock Adversarial examples are not bugs, they are features.
\newblock In {\em Advances in Neural Information Processing Systems (NIPS)},
  2019.

\bibitem{madry2018towards}
Aleksander Madry, Aleksandar Makelov, Ludwig Schmidt, Dimitris Tsipras, and
  Adrian Vladu.
\newblock Towards deep learning models resistant to adversarial attacks.
\newblock In {\em International Conference on Learning Representation (ICLR)},
  2018.

\bibitem{marcus2018deep}
Gary Marcus.
\newblock Deep learning: A critical appraisal.
\newblock 2018.
\newblock arXiv:1801.00631.

\bibitem{nvidiaapex}
NVIDIA.
\newblock Apex - a {P}y{T}orch extension: Tools for easy mixed precision and
  distributed training in {P}y{T}orch.
\newblock \url{https://github.com/NVIDIA/apex}.

\bibitem{papernot2015limitations}
Nicolas Papernot, Patrick McDaniel, Somesh Jha, Matt Fredrikson, Z.~Berkay
  Celik, and Ananthram Swami.
\newblock The limitations of deep learning in adversarial settings.
\newblock In {\em IEEE European symposium on security and privacy (EuroS\&P)},
  2016.

\bibitem{papernot2016distillation}
Nicolas Papernot, Patrick McDaniel, Xi~Wu, Somesh Jha, and Ananthram Swami.
\newblock Distillation as a defense to adversarial perturbations against deep
  neural networks.
\newblock In {\em IEEE symposium on security and privacy (S\&P)}, 2016.

\bibitem{imagenet_benchmark}
{Papers with Code}.
\newblock Image classification on imagenet.
\newblock
  \url{https://paperswithcode.com/sota/image-classification-on-imagenet}.
\newblock Accessed: 2021-06-03.

\bibitem{NEURIPS2019_9015}
Adam Paszke, Sam Gross, Francisco Massa, Adam Lerer, James Bradbury, Gregory
  Chanan, Trevor Killeen, Zeming Lin, Natalia Gimelshein, Luca Antiga, Alban
  Desmaison, Andreas Kopf, Edward Yang, Zachary DeVito, Martin Raison, Alykhan
  Tejani, Sasank Chilamkurthy, Benoit Steiner, Lu~Fang, Junjie Bai, and Soumith
  Chintala.
\newblock {P}y{T}orch: An imperative style, high-performance deep learning
  library.
\newblock In {\em Advances in Neural Information Processing Systems (NIPS)},
  2019.

\bibitem{shafahi2019adversarial}
Ali Shafahi, Mahyar Najibi, Amin Ghiasi, Zheng Xu, John Dickerson, Christoph
  Studer, Larry~S Davis, Gavin Taylor, and Tom Goldstein.
\newblock Adversarial training for free!
\newblock In {\em Advances in Neural Information Processing Systems (NIPS)},
  2019.

\bibitem{szegedy2014intriguing}
Christian Szegedy, Wojciech Zaremba, Ilya Sutskever, Joan Bruna, Dumitru Erhan,
  Ian Goodfellow, and Rob Fergus.
\newblock Intriguing properties of neural networks.
\newblock In {\em International Conference on Learning Representations (ICLR)},
  2014.

\bibitem{forbes}
Rob Toews.
\newblock Deep learning has limits. but its commercial impact has just begun.
\newblock
  \url{https://www.forbes.com/sites/robtoews/2020/02/09/deep-learning-has-limits-but-its-commercial-impact-has-just-begun/?sh=5c38edca6e1a}.
\newblock Accessed: 2021-06-02.

\bibitem{tramer2018ensembletraining}
Florian Tram{\`e}r, Alexey Kurakin, Nicolas Papernot, Ian Goodfellow, Dan
  Boneh, and Patrick McDaniel.
\newblock Ensemble adversarial training: Attacks and defenses.
\newblock In {\em International Conference on Learning Representations (ICLR)},
  2018.

\bibitem{tsipras2018robustness}
Dimitris Tsipras, Shibani Santurkar, Logan Engstrom, Alexander Turner, and
  Aleksander Madry.
\newblock Robustness may be at odds with accuracy.
\newblock In {\em International Conference on Learning Representation (ICLR)},
  2019.

\bibitem{wong2018provable}
Eric Wong and Zico Kolter.
\newblock Provable defenses against adversarial examples via the convex outer
  adversarial polytope.
\newblock In {\em International Conference on Machine Learning (ICML)}, 2018.

\bibitem{wong2019fast}
Eric Wong, Leslie Rice, and J~Zico Kolter.
\newblock Fast is better than free: Revisiting adversarial training.
\newblock In {\em International Conference on Learning Representations (ICLR)},
  2019.

\bibitem{zhang2019stable}
Huan Zhang, Hongge Chen, Chaowei Xiao, Sven Gowal, Robert Stanforth, Bo~Li,
  Duane Boning, and Cho-Jui Hsieh.
\newblock Towards stable and efficient training of verifiably robust neural
  networks.
\newblock In {\em International Conference on Learning Representation (ICLR)},
  2019.

\end{thebibliography}

\clearpage

\appendix
\section{Hyperparameters} \label{app:hyperparams}
    Here we provide the hyperparameters used to train the models in our experiments. Table \ref{tab:hyperparams_baseline} provides the hyperparameters used to adversarially train each model. Table \ref{tab:hyperparams_rrm} provides the hyperparameters used when training a student model using \sys.  Table \ref{tab:hyperparams_prior} provides the hyperparameters used when training a student model using prior works: Robust Data Training (RDT)~\cite{Ilyas2019AdversarialEA}, Knowledge Distillation (KD)~\cite{Goldblum_2020}, Fast Adversarial Training (Fast AT)~\cite{wong2019fast}, and Free Adversarial Training (Free AT)~\cite{shafahi2019adversarial}.
    
\begin{table}[h!]
    \centering
    \caption{Hyperparameters used to train models using standard adversarial training~\cite{madry2018towards} on different datasets.}
    \vspace{0.5em}
    \resizebox{\columnwidth}{!}{%
    \begin{tabular}{@{}lcccc@{}}
    \toprule
    \textbf{Dataset} & \textbf{LR} & \textbf{Batch Size} & \textbf{Epochs} & \textbf{LR Decay} \\
    \midrule
    \cifar & 0.1 & 128 & 150 & 50,100 \\
    \rimagenet & 0.01 & 128 & 150 & 125 \\
    \bottomrule
    \end{tabular}
    }
\label{tab:hyperparams_baseline}
\end{table}

\begin{table}[h!]
    \centering
    \caption{Hyperparameters used to train models using \sys on different datasets.}
    \vspace{0.5em}
    \resizebox{\columnwidth}{!}{%
    \begin{tabular}{@{}lcccc@{}}
    \toprule
    \textbf{Dataset} & \textbf{LR} & \textbf{Batch Size} & \textbf{Epochs} & \textbf{LR Decay} \\
    \midrule
    \cifar & 0.1 & 128 & 48 & cosine \\
    \rimagenet & 0.1 & 128 & 60 & 35,50 \\
    \bottomrule
    \end{tabular}
    }
\label{tab:hyperparams_rrm}
\end{table}

\begin{table}[h!]
    \centering
    \caption{Hyperparameters used to train models using prior works on \cifar .}
    \vspace{0.5em}
    \resizebox{\columnwidth}{!}{%
    \begin{tabular}{@{}lcccc@{}}
    \toprule
    \textbf{Method} & \textbf{LR} & \textbf{Batch Size} & \textbf{Epochs} & \textbf{LR Decay} \\
    \midrule
    RDT~\cite{Ilyas2019AdversarialEA} & 0.1 & 128 & 100 & 65,90 \\
    KD~\cite{Goldblum_2020}  & 0.1 & 128 & 100 & 65,90 \\
    Fast AT~\cite{wong2019fast} & 0.2 (max) & 128 & 40 & cyclic \\
    Free AT~\cite{shafahi2019adversarial} & 0.04 (max) & 128 & 96 & cyclic \\
    \bottomrule
    \end{tabular}
    }
\label{tab:hyperparams_prior}
\vspace{-1em}
\end{table}

\vfill\null
\pagebreak

\section{Penultimate Layer Dimensions} \label{app:pen_dim}
In order to demonstrate that \sys is model-agnostic, we used a variety of classifiers in our experiments. These classifiers belong to different class of architectures (VGG, ResNet \etc) as well as have different penultimate layer dimensions. Since our robust representation loss requires the penultimate layer features of the student and teacher to be of the same dimension, we add an additional fully connected layer after the penultimate layer in certain classifiers. These modifications are summarized in Table~\ref{tab:pen_dim}.

\begin{table}[h!]
    \centering
    \caption{Dimensions of the penultimate layer features of the various classifiers we use in our experiments. To be able to use or robust representation loss, we add an additional fully connected layer to some architectures as specified below.}
    \vspace{0.5em}
    \begin{tabular}{@{}ccc@{}}
    \toprule
    \multirow{2}{*}{\textbf{Classifier}} & \multicolumn{2}{c}{\textbf{Penultimate Layer Dimension}} \\ \cmidrule(lr){2-3}
     & \textbf{Original} & \textbf{After Modification} \\
    \midrule
    \multicolumn{3}{c}{\textbf{\cifar}} \\
    \midrule
    \vggs & 512 & N/A \\
    \vggl & 512 & N/A \\
    \resnets & 512 & N/A \\
    \resnetl & 2048 & 512 \\
    \midrule
    \multicolumn{3}{c}{\textbf{\rimagenet}} \\
    \midrule
    \alexnet & 4096 & 2048 \\
    \vggm & 4096 & 2048 \\
    \resnetl & 2048 & N/A \\
    \bottomrule
    \end{tabular}
\label{tab:pen_dim}
\vspace{-1em}
\end{table}

\vfill\null
\pagebreak

\section{Adversarial Training Results} \label{app:sat_results}
In this section, we include the performance of all the adversarially trained models (SAT~\cite{madry2018towards} and Fast AT~\cite{wong2019fast}) that we used in our experiments. Information regarding the adversary used is provided in parentheses. Refer to Table~\ref{tab:at_results} for these results.

\begin{table}[h!]
    \centering
    \caption{Performance of adversarially trained models we used in our experiments. The AutoPGD attack~\cite{croce2020reliable} was performed using 20 iterations and 5 random restarts.}
    \vspace{0.5em}
    \resizebox{\columnwidth}{!}{%
    \begin{tabular}{@{}ccccc@{}}
    \toprule
    \textbf{Threat Model} & \textbf{Classifier} & \textbf{Method} & \textbf{Natural} & \textbf{AutoPGD} \\
    \midrule
    \multicolumn{5}{c}{\textbf{\cifar}} \\
    \midrule
    \multirow{5}{*}{$\ell_2$, $\epsilon=1.0$} & DNN & SAT & 59.95 & 34.15 \\
     & \vggs & SAT & 78.81 & 46.08 \\
     & \vggl & SAT & 74.64 & 45.52\\
     & \resnets & SAT & 82.81 & 48.99\\
     & \resnetl & SAT & 82.97 & 48.65\\
    \midrule
    \multirow{5}{*}{$\ell_\infty$, $\epsilon=8/255$} & \vggs & Fast AT & 76.94 & 44.11 \\
     & \resnets & Fast AT & 82.60 & 51.30 \\
     & \resnetl & SAT & 85.50 & 49.60 \\
     & \resnetl & Fast AT & 83.73 & 50.65 \\
     & \resnetl & Free AT & 77.74 & 45.41 \\
    \midrule
    \multicolumn{5}{c}{\textbf{\rimagenet}} \\
    \midrule
    \multirow{3}{*}{$\ell_2$, $\epsilon=3.0$} & \alexnet & SAT & 87.67 & 75.24 \\
     & \vggm & SAT & 92.40 & 80.91 \\
     & \resnetl & SAT & 95.47	& 84.36 \\
    \bottomrule
    \end{tabular}
    }
\label{tab:at_results}
\end{table}



\vfill\null
\pagebreak

\section{Training Time Results} \label{app:speedup_results}
The total time required to train a \resnetl model using different methods is reported in this section. For completeness, we report the total training time with teacher overhead included in case of \sys. The total training times on \cifar are reported in Table~\ref{tab:speedup_cifar} and for \rimagenet are reported in Table~\ref{tab:speedup_rimagenet}.

\begin{table}[h!]
    \centering
    \caption{Comparing \sys training time to Free AT~\cite{shafahi2019adversarial} on \cifar using \resnetl. Both methods have been accelerated using DAWNBench improvements~\cite{wong2019fast}. For completeness we provided comparisons when teacher's overhead is taken into account when training models using \sys. Teachers \vggs and \resnets have been trained using Fast AT.}
    \vspace{0.5em}
    \begin{tabular}{@{}lcc@{}}
    \toprule
    \multirow{2}{*}{\textbf{Method}} & \multicolumn{2}{c}{\textbf{Train Time (mins)}} \\ \cmidrule(lr){2-3}
     & \textbf{w/o Teacher} & \textbf{w/ Teacher} \\
    \midrule
    SAT & 1807.58 & - \\
    Fast AT & 192.77 & - \\
    Free AT & 58.44 & - \\
    \midrule
    \sys (\vggs) & 30.22 & 57.91 \\
    \sys (\resnets) & 31.82 & 91.77 \\
    \bottomrule
    \end{tabular}
\label{tab:speedup_cifar}
\end{table}

\begin{table}[h!]
    \centering
    \caption{Comparing \sys training time to SAT~\cite{madry2018towards} on \rimagenet using \vggm and \resnetl. An adversarially trained \alexnet model is used as teacher. For completeness we provided comparisons when teacher's overhead is taken into account when training models using \sys.}
    \vspace{0.5em}
    \begin{tabular}{@{}lcc@{}}
    \toprule
    \multirow{2}{*}{\textbf{Method}} & \multicolumn{2}{c}{\textbf{Train Time (hrs)*}} \\ \cmidrule(lr){2-3}
     & \textbf{w/o Teacher} & \textbf{w/ Teacher} \\
    \midrule
    \multicolumn{3}{c}{\resnetl} \\
    \midrule
    SAT & 253.71 & - \\
    \sys & 13.66 & 46.69 \\
    \midrule
    \multicolumn{3}{c}{\vggm} \\
    \midrule
    SAT & 400.01 & - \\
    \sys & 33.30 & 66.33 \\
    \bottomrule
    \end{tabular}
    \caption*{\textit{*The \resnetl and \vggm models were trained on different machines due to memory constraints.}}
\label{tab:speedup_rimagenet}
\end{table}

\vfill\null
\pagebreak

\section{Artifact Appendix}

%

\subsection{Abstract}

This artifact includes the code necessary to reproduce the experimental results presented in our paper titled ``Transferring Adversarial Robustness Through \syslong''. It is made available in the form of a GitHub repository (final stable URL: \github). Our experiments involve training neural network based image classifiers that are robust against adversarial attacks. Therefore, we provide the necessary training and evaluation scripts, along with all the supporting code. The expected results, as reported in the paper, are : (1)~total training time, and (2)~accuracy of the trained classifier on clean and adversarial test sets. All our code is written in Python.

On the hardware side, the code requires a machine with atleast one GPU with 12~GB memory and storage space >~150~GB to run. We recommend atleast 8~GB of RAM. On the software side, the code requires Python compiler, pip, conda and several other 3rd party Python libraries like PyTorch, IBM's Adversarial Robustness Toolbox, Nvidia's apex \etc Detailed instructions regarding setting up the run-time environment are provided in Section~\ref{sec:install} and the repository README.

\subsection{Artifact check-list (meta-information)}


{\small
\begin{itemize}
  \item {\bf Algorithm: }Our paper presents a novel algorithm called \syslong (\sys). The purpose of this algorithm is to speed up the process of adversarially training neural network based image classifiers.
  \item {\bf Data set: }We perform experiments using two image datasets: \cifar and \rimagenet. The \cifar dataset downloads itself if not available. It requires 341MB storage space. For experiments involving \rimagenet, the full \imagenet dataset needs to be downloaded. Instructions for this are provided in the README of the code repository. It requires 145~GB  storage space.
  \item {\bf Model: }The \cifar experiments are conducted using the following neural networks: \vggs, \vggl, \resnets, \resnetl. The \rimagenet experiments use the following neural networks: \alexnet, \vggm, \resnetl. All the code associated with these networks is provided in the repository. We also make available weights of pre-trained classifiers for quick evaluation.
  \item {\bf Run-time environment: }Our code has been tested on a Linux machine. To prepare the run-time environment, one needs to create a Python virtual environment and install all required Python libraries. The instructions for setting up the run-time environment are provided in Section~\ref{sec:install} and the README in the repository.
  \item {\bf Hardware: } The code requires a machine with atleast one GPU with 12~GB memory and storage space >~150~GB. We recommend running the \rimagenet training scripts on 4~GPUs. Also, we recommend 8~GB of RAM.
  \item {\bf Execution: } Here we provide estimated time taken by different components of our experiments. These estimates were computed on our machine. We ran our experiments on two different machines. The \cifar experiments were run on a machine with an Intel Xenon(R) Gold 6136 CPU, 16~GB RAM, and an Nvidia Titan V GPU. The training scripts took $\sim 5$ hours on average. In total 19 classifiers need to be trained using different methods. The \rimagenet experiments were run on a second machine with an Intel Xenon(R) E5-2690 CPU, 16~GB RAM, and an Nvidia V100 GPU. The training scripts took $\sim 1$ week to run on average. In total 5 classifiers need to be trained using different methods. For both the datasets, the evaluation scripts take $\sim 3$ hours to run in the worst case, with every trained model needing to be evaluated once.
  \item {\bf Metrics: }We report two metrics in our paper: (1)~Training run time and (2)~Accuracy of clean and adversarial test sets. Note that due to differences in hardware, the absolute training times will be different than what is reported in the main paper. However, the speedup (ratio of train times) should be approximately the same.
  \item {\bf Output: }All the expected output will be printed out as stdout on running the evaluation script. The following quantities will be outputted: (1)~average time per training epoch, (2)~its 95\% confidence interval, (3)~total training time, (4)~accuracy on clean test set, and (5)~accuracy on adversarial test set.
  \item {\bf Experiments: } The step-by-step instructions to reproduce the experimental results are provided in the GitHub READMEs. The accuracy numbers will be within a few percentage points of the numbers reported in the main paper. The absolute training time numbers will vary from what is reported in the main paper due to hardware differences. However, the speedup numbers (ratio of training times) will be approximately the same.
  \item {\bf How much disk space required (approximately)?: } 150~GB
  \item {\bf How much time is needed to complete experiments (approximately)?: }On our machine, training all the reported classifiers on \cifar took $\sim 4$ days. Training all the reported \rimagenet classifiers took $\sim 5$ weeks. Evaluating all the classifiers (corresponding to both datasets) took $\sim 3$ days. During reproduction, expect significant variations in these times because of hardware differences.
  \item {\bf Publicly available (explicitly provide evolving version reference)?: } Yes. \url{https://github.com/Ethos-lab/robust-representation-matching}
  \item {\bf Code licenses (if publicly available)?: } MIT License
  \item {\bf Data licenses (if publicly available)?: } \cifar: no license. \imagenet: \url{https://www.image-net.org/download.php}.
\end{itemize}}

\subsection{Description}


\subsubsection{How to access}
Clone GitHub repository, available here (final stable URL): \\ \github

\subsubsection{Hardware dependencies}
The code requires a machine with atleast one GPU with 12~GB memory and storage space >~150~GB. We recommend running the \rimagenet training scripts on 4 GPUs. Also, we recommend 8~GB of RAM.

\subsubsection{Software dependencies}
Our code is written in Python and requires a Python compiler installed along with the python package managers pip and conda. In addition, our code makes use of several 3rd party Python libraries. For instructions regarding how to install all the software dependencies and set up the run-time environment, refer to Section~\ref{sec:install} and the GitHub README.

\subsubsection{Data sets}
We use two datasets in our experiments: \cifar and \rimagenet. \cifar will download itself if not available. For \rimagenet, the entire \imagenet dataset needs to be downloaded. Instructions for this are provided in the GitHub README.

\subsubsection{Models}
The \cifar experiments are conducted using the following neural networks: \vggs, \vggl, \resnets, \resnetl. The \rimagenet experiments use the following neural networks: \alexnet, \vggm, \resnetl. All the code associated with these networks is provided in the repository. We also make available weights of pre-trained classifiers for quick evaluation.

\subsubsection{Security, privacy, and ethical concerns}
All the data we use is publicly available for research. The work presented in our paper introduces no security, privacy, or ethical concerns.

\subsection{Installation} \label{sec:install}

Follow the following steps to set up the run-time enviroment required to run our code:
\begin{enumerate}
    \item Clone the github repository and navigate into it: \\ \verb|git clone https://github.com/pratik18v/| \verb|robust-representation-matching.git &&| \verb|cd robust-representation-matching|
    \item Create a Python virtual environment and activate it: \\ \verb|conda create -n rrm python=3.6 &&| \verb|conda activate rrm|
    \item Install dependencies: \\ \verb|pip install -r| \verb|requirements.txt|
    \item Install apex using instructions here: \\ \url{https://github.com/NVIDIA/apex#quick-start}
\end{enumerate}
All the instructions to setup the run-time enviroment are also provided in the GitHub README.



\subsection{Evaluation and expected results}

We demonstrate that our proposed algorithm (\sys) trains adversarially robust image classifiers faster than previous state-of-the-art method, at the same time attaining better robustness. For this, we train neural networks using several prior methods and compare them to our method. We perform comparison using two metrics: (1)~total training time, and (2)~accuracy on clean and adversarial test sets. We show that our method has the lowest total training time. Compared to the previous fastest method, our method trains classifier with higher adversarial accuracy. The accuracy numbers can be reproduced within a few percentage points of the numbers reported in the main paper. The absolute training time numbers will vary from what is reported in the main paper due to hardware differences. However, the speedup numbers (ratio of training times) can be reproduced to a value approximately similar to the reported value. The detailed steps to reproduce our results are laid out in the READMEs available in our GitHub repository.



\subsection{Version}
Based on the LaTeX template for Artifact Evaluation V20220119.

\end{document}